%% file: main.tex
\pdfoutput=1
\documentclass[journal]{IEEEtran} 
\usepackage[pdftex]{graphicx}
\graphicspath{{../pdf/}{../jpeg/}}
\DeclareGraphicsExtensions{.pdf,.jpeg,.png}
\usepackage{algorithmic}
\usepackage{fancyhdr}
\usepackage{array}
\usepackage{floatrow}
\usepackage{fixltx2e}
\usepackage{stfloats}
\usepackage{url}
\usepackage{times}
\usepackage{color}
\usepackage{verbatim}
\usepackage{hyperref}
\usepackage{subfiles}
\usepackage{caption}
\usepackage{subcaption}
\usepackage{lipsum}
\usepackage{xr}
\usepackage{nameref}
\usepackage{zref-xr}
\usepackage{adjustbox}
\usepackage{amsmath, amssymb}
\usepackage[ansinew]{inputenc}
\usepackage{mathtools}
\usepackage{esvect}
\usepackage{xspace}
\usepackage{standalone}
\usepackage{fixltx2e}
\usepackage{authblk}
\usepackage{xcolor}
\usepackage{booktabs}
\usepackage{multirow}
\usepackage{cleveref}
\usepackage{tabularx}
\usepackage{gensymb}
\usepackage{pifont}

\newcommand{\xmark}{\ding{55}}%
\newcommand{\airl}{Air~Learning\xspace}

\newcommand{\ras}[0]{Ras-Pi~4\xspace}
\newcommand{\Fig}[1]{Figure~\ref{#1}}
\newcommand{\Tab}[1]{Table~\ref{#1}}

\newcommand{\zoneD}{\textit{Zone~3}\xspace}

\newcommand{\noobj}{\textit{No~Obstacles}\xspace}
\newcommand{\staticobj}{\textit{Static~Obstacles}\xspace}
\newcommand{\dynamicobj}{\textit{Dynamic~Obstacles}\xspace}

\hypersetup{
    colorlinks=true,
    linkcolor=red,
    filecolor=magenta,      
    urlcolor=cyan,
}

\newcommand{\blue}[1]{\textcolor{black}{#1}}

\fancypagestyle{firstpage}{
  \fancyhf{}
  
  \fancyhead[C]{\vspace{5pt}\normalsize{To Appear in Springer Machine Learning Journal (Special Issue on Reinforcement for Real Life). \\ Preprint Version. Accepted April, 2021.
      \vspace{10pt}}} 
  \fancyfoot[C]{\thepage}
}
\title{Air Learning: A Deep Reinforcement Learning Gym for Autonomous Aerial Robot Visual Navigation}

\renewcommand{\baselinestretch}{1.0}
\begin{document}

\author[$\dagger$]{Srivatsan Krishnan}
\author[$\ddagger$]{Behzad Boroujerdian}
\author[$\dagger$]{William Fu}
\author[$\mp$]{Aleksandra Faust}
\author[$\dagger$$\ddagger$]{Vijay Janapa Reddi}
\affil[ ]{$^\dagger$ Harvard University,   $^\ddagger$The University of Texas at Austin,   $^\mp$ Google Research}
\affil[ ]{\texttt{\url{https://github.com/harvard-edge/airlearning}}}

\maketitle

\thispagestyle{firstpage}
\pagestyle{empty}

\newcommand{\blueDebug}[1]{}
\input{abstract.tex}
\input{intro}
\input{challenges.tex}
\input{related-work.tex}

\input{infrastructure}
\input{prelude.tex}

\input{policy_exploration}
\input{policy-eval.tex}

\input{sys-eval.tex}

\input{mitigation}
\input{future-work.tex}
\input{conclusion}
\input{ack}

\bibliographystyle{ieeetr}
\bibliography{references}
\input{appendix.tex}

\end{document}

%% file: abstract.tex
\begin{abstract}
We introduce \airl, an open-source simulator, and a gym environment for deep reinforcement learning research on resource-constrained aerial robots. Equipped with domain randomization, Air Learning exposes a UAV agent to a diverse set of challenging scenarios. We seed the toolset with point-to-point obstacle avoidance tasks in three different environments and Deep Q Networks (DQN) and Proximal Policy Optimization (PPO) trainers.  \airl assesses the policies' performance under various quality-of-flight (QoF) metrics, such as the energy consumed, endurance, and the average trajectory length, on resource-constrained embedded platforms like a Raspberry Pi. We find that the trajectories on an embedded Ras-Pi are vastly different from those predicted on a high-end desktop system, resulting in up to $40\%$ longer trajectories in one of the environments. To understand the source of such discrepancies, we use \airl to artificially degrade high-end desktop performance to mimic what happens on a low-end embedded system. We then propose a mitigation technique that uses the hardware-in-the-loop to determine the latency distribution of running the policy on the target platform (onboard compute on aerial robot). A randomly sampled latency from the latency distribution is then added as an artificial delay within the training loop. Training the policy with artificial delays allows us to minimize the hardware gap (discrepancy in the flight time metric reduced from 37.73\% to 0.5\%). Thus, Air Learning with hardware-in-the-loop characterizes those differences and exposes how the onboard compute's choice affects the aerial robot's performance. We also conduct reliability studies to assess the effect of sensor failures on the learned policies. All put together, \airl enables a broad class of deep RL research on UAVs. The source code is available at:~\texttt{\url{http://bit.ly/2JNAVb6}}.
\end{abstract}

%% file: intro.tex
\section{Introduction}
Deep Reinforcement Learning (DRL) has shown promising results in domains like sensorimotor control for cars~\cite{nvidia-car-deep-learning}, indoor robots~\cite{autorl}, as well as UAVs~\cite{rl-drones-levine,learning-to-crash}. Deep RL's ability to adapt and learn with minimum apriori knowledge makes them attractive for use in complex systems~\cite{rl_control_thesis}.

Unmanned Aerial Vehicles (UAVs) serve as a great platform for advancing state of the art for DRL research. UAVs have practical applications, such as search and rescue~\cite{search-rescue}, package delivery~\cite{package-delivery,uav-cargo-delivery}, construction inspection~\cite{inspection}. Compared to other robots such as self-driving cars, robotic arm, they are vastly cheap to prototype and build, which makes them truly scalable.\footnote{For example, an FPV hobbyist drone can be built under \$100:~\url{https://bit.ly/2TR3rMQ}} Also, UAVs have fairly diverse control requirements. Targeting low-level UAV control (e.g. attitude control) requires continuous control (e.g., angular velocities) whereas, targeting high-level tasks such as point-to-point navigation can use discrete control. Last but not least, at deployment time they must be a fully autonomous system, running onboard  computationally- and energy-constrained computing hardware.

\begin{figure}[t!]
\centering
        \includegraphics[width=0.95\columnwidth,keepaspectratio]{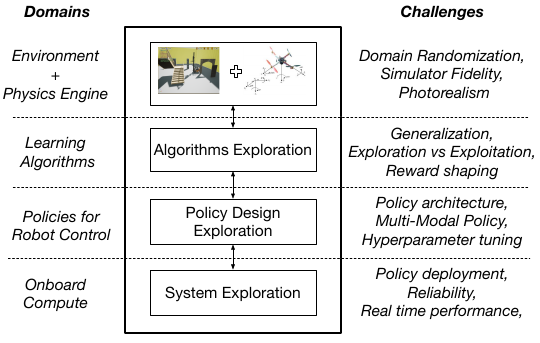}
        \caption{\small Aerial robotics is a cross-layer, interdisciplinary field. Designing an autonomous aerial robot to perform a task involves interactions between various boundaries, spanning from environment modeling down to the choice of hardware for the onboard compute.}
    \label{fig:airlearning-motivation}
\end{figure}

But despite the promise of Deep RL, there are several practical challenges in adopting reinforcement learning for the UAV navigation task as shown in \Fig{fig:airlearning-motivation}. Broadly, the problems can be grouped into four main categories: (1) environment simulator, (2) learning algorithms, (3) policy architecture, and (4) deployment on resource-constrained UAVs. To address these challenges, the boundaries between reinforcement learning algorithms, robotics control, and the underlying hardware must soften. The figure illustrates the cross-layer, and interdisciplinary nature of the field, spanning from environment modeling to the underlying system. Each layer, in isolation, has a complex design space that needs to be explored for optimization. In addition, there are interactions across the layers that are also important to consider (e.g., policy size on a power-constrained mobile or embedded computing system). Hence, there is a need for a platform that can aid interdisciplinary research. More specifically, we need a research platform that can benchmark each of the layers individually (for depth), as well as end-to-end execution for capturing the interactions across the layers (for breadth).

In this paper, we present \airl (Section~\ref{sec:infrastructure}) ---an open source Deep RL research simulation suite and benchmark for autonomous UAVs. As a simulation suite of tools, \airl\ provides a scalable and cost-effective applied reinforcement learning research. It augments existing frameworks such as AirSim~\cite{AirSim} with capabilities that make it suitable 
for Deep RL experimentation. \emph{As a gym, \airl\ enables RL research for resource constrained systems.}

\airl addresses the simulator level challenge, by providing domain randomization. We develop a configurable environment generator with a range of knobs to generate different environments with varying difficulty levels. The knobs (randomly) tune the number of static and dynamic obstacles, their speed (if relevant), texture and color, arena size, etc. In the context of our benchmarking autonomous UAV navigation task, the knobs help the learning algorithms' generalize well without overfiting to a specific instance of an environment.\footnote{The environment generator can be applied to other challenges in aerial robots, such as detecting thin wires and coping with translucent objects.}

\airl\ addresses the learning challenges (RL algorithm, policy design, and reward optimization) by exposing the environment generator as an OpenAI gym~\cite{openai} interface and integrating it with Baselines~\cite{stable-baselines}, which has high-quality implementations of the state-of-the-art RL algorithms. We provide templates which the researchers can use for building multi-modal input policies based on Keras/Tensorflow. And as a DRL benchmark, the OpenAI gym interface enables easy addition of new deep RL algorithms.
At the point of writing this paper, we provide two different reinforcement learning algorithms Deep Q-Networks (DQN) \cite{dqn} and Proximal Policy Optimization (PPO) \cite{ppo-paper}. DQN is offpolicy, discrete action RL algorithm, and PPO is onpolicy, continuous action control of UAVs. Both come ready with curriculum learning~\cite{bengio2009curriculum} support.

To address the resource-constrain challenge early on in the design and development of Deep RL algorithms and policies, \airl uses a ``hardware-in-the-loop'' (HIL)~\cite{HIL} method to enable robust hardware evaluation without risking real UAV platform. Hardware in the loop, which requires plugging in the computing platform used in the UAV into the software simulation, is a form of real-time simulation that allows us to understand how the UAV responds to simulated stimuli on a target hardware platform.\footnote{Demonstration of HIL:~\url{https://bit.ly/2NDRjex}.} HIL simulation helps us quantify the real-time performance of reinforcement learning policies on various compute platforms, without risking experiments on real robot platforms before they are ready. 

We use HIL simulation to understand how a policy performs on an embedded compute platform that might potentially be the onboard computer of the UAVs. To enable systematic HIL evaluation, we use a variety of Quality-of-Flight (QoF) metrics, such as the total energy consumed by the UAV, the average length of its trajectory and endurance, to compare the different reinforcement learning policies. To demonstrate that \airl's HIL simulation is essential and that it can reveal interesting insights, we take the best performing policy from our policy exploration stage and evaluate the performance of the policy on a resource-constrained low-performance platform (\ras) and compare it with a high-performance desktop counterpart (Intel Core-i9). The difference between the \ras and the Core-i9 based performance for the policy is startling. The \ras sometimes takes trajectories that are nearly \textit{40\%} longer in some environments.  We investigate the reason for the difference in the performance of the policy on \ras versus Intel Core-i9 and show that the choice of onboard compute platform directly affects the policy processing latency, and hence the trajectory lengths. The discrepancy in the policy behavior from training to deployment hardware is a challenge that must be taken into account when designing the DRL algorithm for a resource-constrained robot. We define this behavior as `Hardware induced gap' because of the performance gap in training machine versus deployment machine.
We use a variety of metrics to quantify the hardware gap, such as percentage change between the QoF metrics that include flight time, success rate, the energy of flight, and trajectory distance.

In summary, we present an open-source gym environment and research platform for Deep RL research for autonomous aerial vehicles. The contributions within this context include:

\begin{itemize}
    \item We present an open source benchmark to develop and train different RL algorithms, policies, and reward optimizations using regular and curriculum learning.
    
    \item We present a UAV mapless navigation task benchmark for RL research on resource constrained systems. 
    
    \item We present a random environment generator for domain randomization to enable RL generalization.
   
    \item We introduce and show `Hardware induced gap' -- that the policy's behavior depends on a computing platform it is running on, and that the same policy can result in a very different behavior if the target deployment platform is very different from than the training platform.
    \item We describe the significance of taking energy consumption and the platform's processing capabilities into account when evaluating policy success rates. 
    \item{To alleviate the hardware-induced gap, we train a policy using HIL to match the target platform's latencies. Using this mitigation technique, we minimized the hardware gap between the training platform and resource-constrained target platform from 38\% to less than 0.5\% on flight time, 16.03\% to 1.37\% on the trajectory length, and 15.49\% to 0.1\% on the energy of flight metric.}
    
\end{itemize}

\airl\ will be of interest to both fundamental and applied RL research community. The point to point UAV navigation benchmark can yield to progress on fundamental RL algorithm development for resource-constrained systems where training and deployment platforms are different. From that point of view, \airl\ is another OpenAI Gym environment. For the applied RL researchers, interested in RL applications for UAV domains such as source seeking, search and reuse, etc., \airl\ serves as a simulation platform and toolset with for full-stack research and development. 

%% file: challenges.tex
\section{Real World Challenges}
\label{sec:challenges}
We describe the real-world challenges associated with developing Deep RL algorithms on resource-constrained UAVs. We consolidate the challenges into four categories, namely Environment simulator, challenges related to the learning algorithm, policy selection challenges, and hardware-level challenges.

\textit{Environment Simulator Challenges:} The first challenge is that Deep RL algorithms targetted for robotics need simulator. Collecting large amounts of real-world data is challenging because most commercial and off-the-shelf UAVs operate for less than 30~mins. To put this into perspective, creating a dataset as large as the latest ``ImageNet'' by Tencent for ML Images~\cite{tencent-images} would take close to 8000 flights (assuming a standard 30 FPS camera), thus making it a logistically challenging issue. But perhaps an even more critical and difficult aspect of this data collection is that there is a need for negative experiences, such as obstacle collisions, which can severely drive up the cost and logistics of collecting  data~\cite{learning-to-crash}. More importantly, it has been shown the environment simulator having high fidelity and ability to perform domain randomization aids in the better generalization of reinforcement learning algorithms~\cite{domain-rand}. Hence,  any infrastructure for Deep RL must have features to address these challenges to deploy RL policies in real-world robotics applications. 

\textit{Learning Algorithm Challenges:} The second challenge is associated with reinforcement learning algorithms. Choosing the right variant of a reinforcement learning algorithm for a given task requires fairly exhaustive exploration. Furthermore, since the performance and efficiency of a particular reinforcement learning algorithm are greatly influenced by its reward function, to get good performance, there is a need to perform design exploration between the reinforcement learning algorithms and its reward function. Though these challenges are innate to the Deep RL domain, having an environment simulator exposed as a simple interface~\cite{openai} can allow us to efficiently automate the RL algorithm selections, rewards shaping, hyperparameter tuning~\cite{auto-rl}.

\textit{Policy Selection Challenges:} The third challenge is associated with the selection of policies for robot control. Choosing the right policy architecture is a fairly exhaustive task. Depending upon the available sensor suite on the robot, the policy can be uni-modal or multi-modal in nature. Also, for effective learning, the hyperparameters associated with the policy architecture have to be appropriately tuned. Hyperparameter tuning and policy architecture search is still an active area of research, which has lead to techniques such as AutoML~\cite{auto-ml} to determine the optimal neural network architecture. In the context of DRL policy selection, having a standard machine learning back-end tool such as Tensorflow/Keras~\cite{tensorflow} can allow DRL researchers (or roboticist) to automate the policy architecture search.

\textit{Hardware-level Challenges:} The fourth challenge is regarding the deployment of Deep RL policies on the resource-constrained UAVs. Since UAVs are mobile machines, they need to accomplish their tasks with a limited amount of onboard energy. Because onboard compute is a scarce resource and RL policies are computationally intensive, we need to carefully co-design the policies with the underlying hardware so that the compute platform can meet the real-time requirements under power constraints. As the UAV size decreases, the problem exacerbates because battery capacity (i.e., size) decreases, which reduces the total onboard energy (even though the level of intelligence required remains the same). For instance, a nano-UAV such as a CrazyFlie~\cite{bitcraze} must have the same autonomous navigation capabilities as compared to its larger mini counterpart, e.g., DJI-Mavic Pro~\cite{DJI} while the CrazyFlie's onboard energy is $\frac{1}{15}$th that of the Mavic Pro. Typically in Deep RL research for robotics, the system and onboard computers are based on commercial off the shelf hardware platforms. However, whether the selection of these compute platforms is optimal is mostly unknown. Hence, having the ability to characterize the onboard computing platform early on can lead to resource-friendly DRL policies.

\airl is built with features to overcome the challenges listed above. Due to the interdisciplinary nature of the tool, it provides flexibility to researchers to focus on a given layer (e.g., policy architecture design) while also understanding its impact on the subsequent layer (e.g., hardware performance). In the next section, we describe the related work and list of features that \airl supports out of the box.

%% file: related-work.tex
\section{Related Work}
\label{sec:related-work}
Related work in Deep RL toolset and benchmarks can be divided into three categories. The first category of related work includes environments for designing and benchmarking new DRL algorithms. In the second category of related work includes tools used specifically for DRL based aerial robots. In the third category, we include other learning-based toolsets that support features that are important for Deep RL training. The feature list and comparison of related work to \airl are tabulated in Table~\ref{tab:comparison}.

\textbf{Benchmarking Environments:} The first category of related work includes benchmarking environments such as OpenAI Gym~\cite{openai}, Arcade Learning Environments~\cite{atari_arcade}, and MujoCo~\cite{MuJoCo}. These environments are simple by design and allow designing and benchmarking of new Deep RL algorithms. However, using these environments for real-life applications such as robotics is challenging because they do not address the hardware-level challenges (Section~\ref{sec:challenges}) for transferring trained RL policies to real robots. Air Learning addresses these limitations by introducing Hardware-in-the-Loop (HIL), which allows end-user to benchmark and characterize the RL policy performance on a given onboard computing platform.

 \textbf{UAV Specific Deep RL Benchmarks:} The second category of related work includes benchmarks that focus on UAVs. For example, AirSim~\cite{AirSim} provides a high-fidelity simulation and dynamics for the UAVs in the form of a plugin that can be imported in any UE4 (Unreal Engine~4)~\cite{unreal} project. However, there are three AirSim limitations that AirLeaning addresses. First, the generation of the environment that includes domain randomization for the UAV task is left to the end-user to either develop or source it from the UE4 market place. The domain randomizations~\cite{domain-rand} are very critical for generalization of the learning algorithm, and we address this limitation in AirSim using the Air Learning environment generator.

Second, AirSim does not model UAV energy consumption. Energy is a scarce resource in UAVs that affects overall mission capability. Hence, learning algorithms need to be evaluated for energy efficiency. Air Learning uses energy model~\cite{mav-bench} within AirSim to evaluate learned policies. Air Learning also allows studying the impact of the performance of the onboard compute platform on the overall energy of UAVs, allowing us to estimate in the simulation how many missions UAV can do, without running in the simulation.

Third, AirSim does not offer interfaces with OpenAI gym or other reinforcement learning framework such as stable baselines\cite{stable-baselines}. We address this drawback by exposing the Air Learning random environment generator with OpenAI gym interfaces and integrate it with a high-quality implementation of reinforcement learning algorithms available in the framework such as baselines~\cite{stable-baselines} and Keras-RL~\cite{keras-rl}. Using Air Learning, we can quickly explore and evaluate different RL algorithms for various UAV tasks.

\begin{table*}[]
\resizebox{\columnwidth}{!}{
\begin{tabular}{|l|l|l|l|l|l|l|l|l|l|}
\hline
\multicolumn{1}{|c|}{\multirow{2}{*}{\textbf{Features}}} &
  \multicolumn{3}{c|}{\textbf{UAV Specific}} &
  \multicolumn{6}{c|}{\textbf{UAV Agnostic}} \\ \cline{2-10} 
\multicolumn{1}{|c|}{} &
  \textbf{Air Learning} &
  \textbf{AirSim} &
  \textbf{GymFC} &
  \textbf{CARLA} &
  \textbf{Gazebo} &
  \textbf{PyRobot} &
  \textbf{Robot-Grasping} &
  \textbf{\blue{ROBEL}~\cite{robel}} &
  \textbf{\blue{SenseAct}~\cite{kindered}} \\ \hline
Photorealism                  & \checkmark & \checkmark & \xmark & \checkmark & \xmark & \xmark & \checkmark  & \xmark & \xmark \\ \hline
Domain Randomization          & \checkmark & \xmark & \xmark & \checkmark & \xmark & \checkmark & \checkmark & \xmark & \xmark \\ \hline
Open-AI Gym Interface         & \checkmark & \xmark & \checkmark & \checkmark & \checkmark & \checkmark & \checkmark & \checkmark & \checkmark  \\ \hline
RL Algorithm Exploration      & \checkmark & \checkmark & \checkmark & \checkmark & \checkmark  & \checkmark & \checkmark & \checkmark & \checkmark  \\ \hline
ML Backend Integration        & \checkmark & \checkmark & \checkmark & \checkmark & \checkmark & \checkmark & \checkmark  & \checkmark & \checkmark \\ \hline
UAV Physics                   & \checkmark & \checkmark & \checkmark & \xmark & \checkmark & \xmark &  \xmark  & \xmark & \xmark\\ \hline
Energy Modelling              & \checkmark & \xmark & \xmark & \xmark & \xmark & \xmark & \xmark  & \xmark & \xmark  \\ \hline
Compute Benchmarking          & \checkmark & \xmark & \xmark & \xmark & \xmark & \xmark & \xmark & \checkmark & \checkmark \\ \hline
RL policy Deployment on Robot & \checkmark & \checkmark & \checkmark & \checkmark & \checkmark & \checkmark & \checkmark & \checkmark & \checkmark  \\ \hline
\end{tabular}}
\caption{Comparison of features commonly present in Deep RL research infrastructures. \checkmark~denotes that the feature exists. \xmark~denotes missing feature or requires significant effort from end-user to enable that feature.}
\label{tab:comparison}
\end{table*}

Another related work that uses a simulator and OpenAI gym interface in the context of UAVs is GYMFC~\cite{gym-fc}. GYMFC uses Gazebo~\cite{gazebo} simulator and OpenAI gym interfaces for training an attitude controller for UAVs using reinforcement learning. The work primarily focuses on replacing the conventional flight controller with a real-time controller based on a neural network. This is a highly specific, low-level task. 
We focus more on high-level tasks, such as point-to-point UAV navigation in an environment with static and dynamic obstacles, and we provide the necessary infrastructure to carry research to enable on-edge autonomous navigation in UAVs. Adapting this work to support a high-level task such as navigation will involve overcoming the limitations of Gazebo, specifically in the context of photorealism. One of the motivations of building AirSim is to overcome the limitations of Gazebo by using state-of-the-art rendering techniques for modeling the environment, which is achieved using robust game engines such as Unreal~Engine~4~\cite{unreal} and Unity~\cite{unity}. 

\textbf{UAV Agnostic Deep RL Benchmarks:} The third category of related work includes Deep RL benchmarks used for other robot tasks, such as grasping by a robotic arm or self-driving car. These related work are highly relevant to \airl because it contains essential features that improve the utility/performance of Deep RL algorithms.

The most prominent work in learning-based approaches for self-driving cars is CARLA~\cite{carla}. It supports a photorealistic environment built on top of a game engine. It also exposes the environment as an OpenAI gym interface, which allows researchers to experiment with different Deep RL algorithms. The physics is based on the game engine, and they do not model energy or focus on the compute hardware performance. Since the CARLA was built explicitly for self-driving cars, porting these features to UAVs will require significant engineering effort.

For the robotic arm grasping/manipulation task, prior work~\cite{not-qt-opt,qt-opt,robel,gu_off_policy} include infrastructure support to train and deploy Deep RL algorithms on these robots. In ~\cite{collective-learning}, they introduce collective learning where they provide distributed infrastructure to collect large amounts of data with real platform experiments. They introduce an asynchronous variant of guided policy search to maximize the utilization (computer and synchronization between different agents), where each agent trains a local policy while a single global policy is trained based on the data collected from individual agents. 
However, these kinds of robots are fixed in a place; hence, they are not limited by energy or by onboard compute capability. So the inability to process or calculate the policy's outcome in real-time only slows down the grasping rate. It does not cause instability. In UAVs, which have a higher control loop rate, uncertainty due to slow processing latency can cause fatal crashes~\cite{scenario_1_2,scenario_1_1}.

For \blue{mobile robots} with/without grasping such as LocoBot~\cite{Locobot}, PyRobot~\cite{pyrobot2019}, ROBEL~\cite{robel} provides open-source tools and benchmarks for training and deploying Deep RL policies on the LocoBot. The simulation infrastructure is based on Gazebo or MuJoCo, and hence it lacks photorealism in the environment and other domain randomization features. Similar to CARLA and robot grasping benchmarks, PyRobot does not model energy or focus on computing hardware performance.

In softlearning~\cite{sac}, the authors apply a soft-actor critic algorithm for the quadrupedal robot. They use Nvidia TX2 on the robot for data collection and also running the policy. The data collected is then used to train the global policy, which is then periodically updated to the robot. In contrast, in our work, we show that training policy on a high-end machine can result in a discrepancy in performance for aerial robot platform. Aerial robots are much more complex to control and unstable compared to ground-based quadrupedal robots. Hence small differences in processing time can hinder its safety. We propose training a policy using the HIL technique with the target platform's latency distribution to mitigate the difference.

\begin{figure*}[h!]
\centering
  \includegraphics[width=0.98\linewidth]{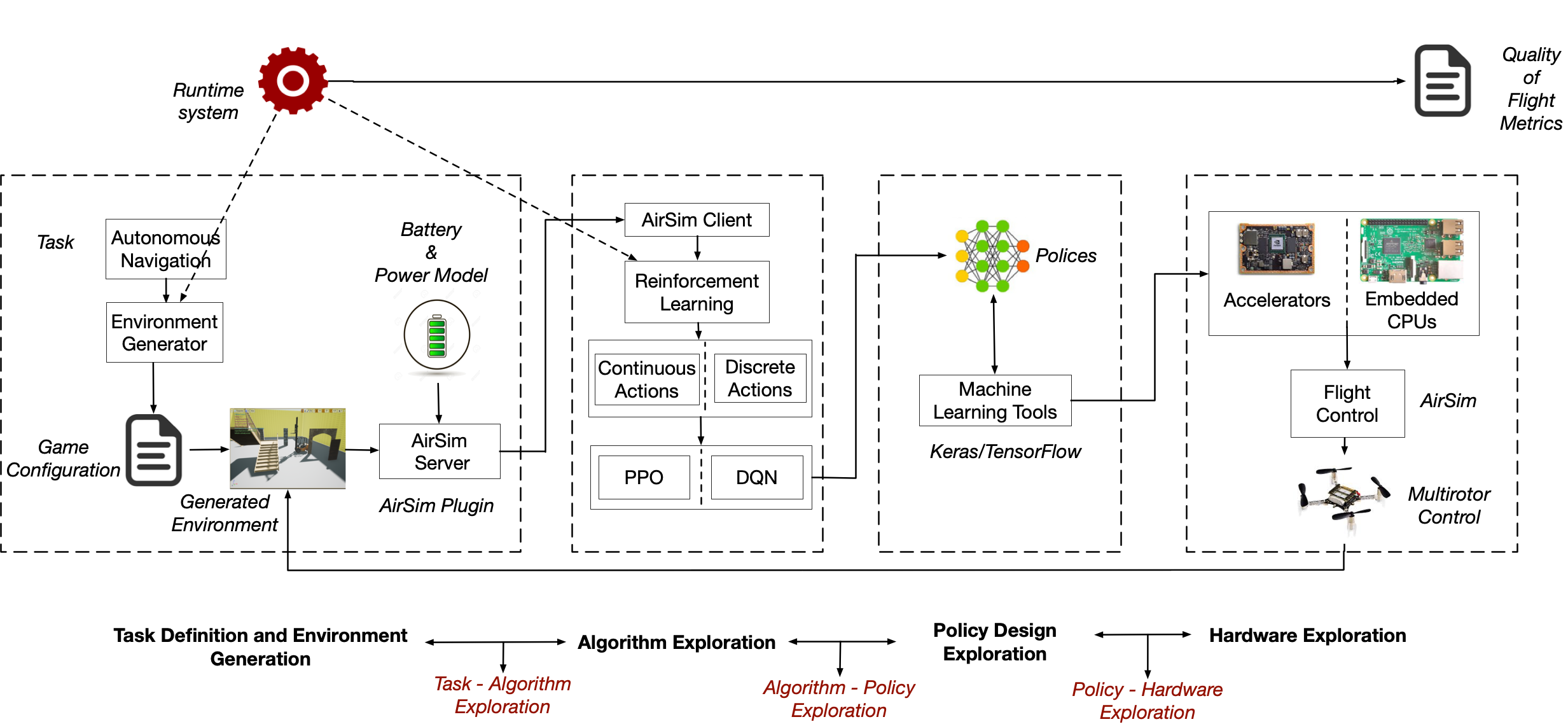}
  \caption{\airl toolset for Deep RL benchmarking in autonomous aerial machines. Our toolset consists of four main components. First, it has a configurable random environment generator built on top of UE4, a photo-realistic game engine that can be used to create a variety of different randomized environments. Second, the random environment generators are integrated with AirSim, OpenAI gym, and baselines for agile development and prototyping different state of the art reinforcement learning algorithms and policies for autonomous aerial vehicles. Third, its backend uses tools like Keras/Tensorflow that allow the design and exploration of different policies. Lastly, \airl uses the ``hardware in the loop'' methodology for characterizing the performance of the learned policies on real embedded hardware platforms. In short, it is an interdisciplinary tool that allows researchers to work from algorithm to hardware with the intent of enabling intra- and inter-layer understanding of execution. It also outputs a set of ``Quality-of-Flight'' metrics to understand execution.}
  \label{fig:airlearning-infrastructure}
\end{figure*}

\textbf{Effect of Action Time in RL Agents:}
Prior works~\cite{reactive-rl,control-delays} have studied the relationship between decision making time (i.e., time taken to decide an action) and task performance in RL agents. The authors propose reactive reinforcement learning algorithms propose a new ``reactive SARSA'' algorithm that orders computational components without affecting the training convergence to make decision making faster. In Air Learning, we expose a similar effect where differences in training hardware (high-end CPU/GPU) and deployment hardware (embedded CPUs) can result in entirely different agent behavior. To that end, we propose a novel action scaling technique based on Hardware-in-the-loop that minimizes the differences between training and deployment of the agent on resource-constrained hardware. Unlike ``reactive SARSA''~\cite{reactive-rl}, we do not make any changes the RL algorithm.

Another related work~\cite{control_delays_2} studies the impact of delays in the action time in the robotic arm's context. The authors use previously computed action until a new action is processed. We study the same problem in aerial robots, where we show that the differences in training and deployment hardware are another source of introducing processing delays and often overlooked. Since drones are deployed in a more dynamic environment, delayed action reduces the drones' reactivity and can severely hinder their safety. To mitigate the performance gaps (hardware gap), we use the HIL methodology to model the target hardware delays and use them for training the policy.

In summary, \airl provides an open source toolset and benchmark loaded with the features to develop Deep RL based applications for UAVs. It helps design effective policies, and also characterize them on an onboard computer using the HIL methodology and quality-of-flight metrics. With that in mind, it is possible to start optimizing algorithms for UAVs, treating the entire UAV and its operation as a system.

%% file: infrastructure.tex
\section{\airl}
\label{sec:infrastructure}

In this section, we describe the various \airl components. The different stages are shown in \Fig{fig:airlearning-infrastructure}, which allows researchers to develop and benchmark learning algorithms for autonomous UAVs. \airl consists of six keys components: an  environment generator, an algorithm exploration framework, closed-loop real-time hardware in the loop setup, an energy and power model for UAVs, quality of flight metrics that are conscious of the UAV's resource constraints, and a runtime system that orchestrates all of these components. By using all these components in unison, \airl allows us to fine-tune algorithms for the underlying hardware carefully.

\subsection{Environment Generator}
\label{sec:env-gen}

Learning algorithms are data hungry, and the availability of high-quality data is vital for the learning process. Also, an environment that is good to learn from should include different scenarios that are challenging for the robot. By adding these challenging situations, they learn to solve those challenges. For instance, for teaching a robot to navigate obstacles, the data set should have a wide variety of obstacles (materials, textures, speeds, etc.) during the training process.

We designed an environment generator specifically targeted for autonomous UAVs. \airl's environment generator creates high fidelity photo-realistic environments for the UAVs to fly in. The environment generator is built on top of UE4 and uses the AirSim UE4~\cite{AirSim} plugin for the UAV model and flight physics. The environment generator with the AirSim plugin is exposed as OpenAI gym interface. 

The environment generator has different configuration knobs for generating challenging environments. The configuration knobs available in the current version can be classified into two categories. The first category includes the parameters that can be controlled via a game configuration file. The second category consists of the parameters that can be controlled outside the game configuration file. The full list of parameters that can be controlled are shown in tabulated in \Tab{tab:game-config}. For more information on these parameters, please refer appendix.

\begin{table*}[]
\centering
\renewcommand*{\arraystretch}{1}
\resizebox{\textwidth}{!}{  
\begin{tabular}{|l|c|l|}
\hline
\textbf{Parameter} 
& \textbf{Format}
& \textbf{Description} 
\\ \hline \hline
\textit{Arena Size }
& {\texttt{[}}\texttt{length}, \texttt{width}, \texttt{height}{\texttt{]}} 
 
& Spawns a rectangular arena of ``{length}" x ``{width}" x ``{height}".                       \\ \hline
\textit{Wall Colors}          & {\texttt{[}}\texttt{R}, \texttt{G}, \texttt{B}{\texttt{]}}                                      & The colors of the wall of in [Red, Green, Blue] color format.                                                   \\ \hline
\textit{Asset}                & \texttt{<folder name>}                                                        & \airl allows any UE4 asset to be imported into the project.                                                                                                             \\ \hline
\textit{\# Static Obstacles}  & Scalar Integer                                             & The number of static obstacles in the arena.                                                                                                                                                                                                   \\ \hline
\textit{\# Dynamic Obstacles} & Scalar Integer                                            & The number of the dynamic obstacle in the arena.                                                                                                                                                                                                       \\ \hline
\textit{Seed}                 & Scalar Integer                                             & Seed value used in randomization.                                                                                                                                                                            \\ \hline
\textit{Minimum Distance}     & Scalar Integer                                             & Minimum distance between two obstacle in the arena.                                                                                                                                                                                                                                                   \\ \hline
\textit{Goal Position}        & {\texttt{[}}\texttt{X}, \texttt{Y}, \texttt{Z}{\texttt{]}}                                       & Sets the goal position in X, Y and Z coordinates.                                                                                                                                                                                                          \\ \hline
\textit{Velocity}             & {\texttt{[}}\texttt{V$_{max}$}, \texttt{V$_{min}$}{\texttt{]}}                             & Velocity of the dynamic obstacle between V$_{max}$ and V$_{min}$ .                                                                                                                                                                                                                                                                                   \\  \hline
\textit{Materials}            & \texttt{<folder name>}                                                        & Any UE4 material can be assigned to the UE4 asset.  \\ \hline
\textit{Textures}             & \texttt{<folder name>}                                                         & Any UE4 Texture can be assigned to the UE4 asset. \\  \hline
\end{tabular}
}\caption{List of configurations available in current version of \airl environment generator.}
\label{tab:game-config}
\end{table*}

\begin{figure*}[t!]
    \begin{subfigure}{0.3\linewidth}
        \includegraphics[width=\textwidth]{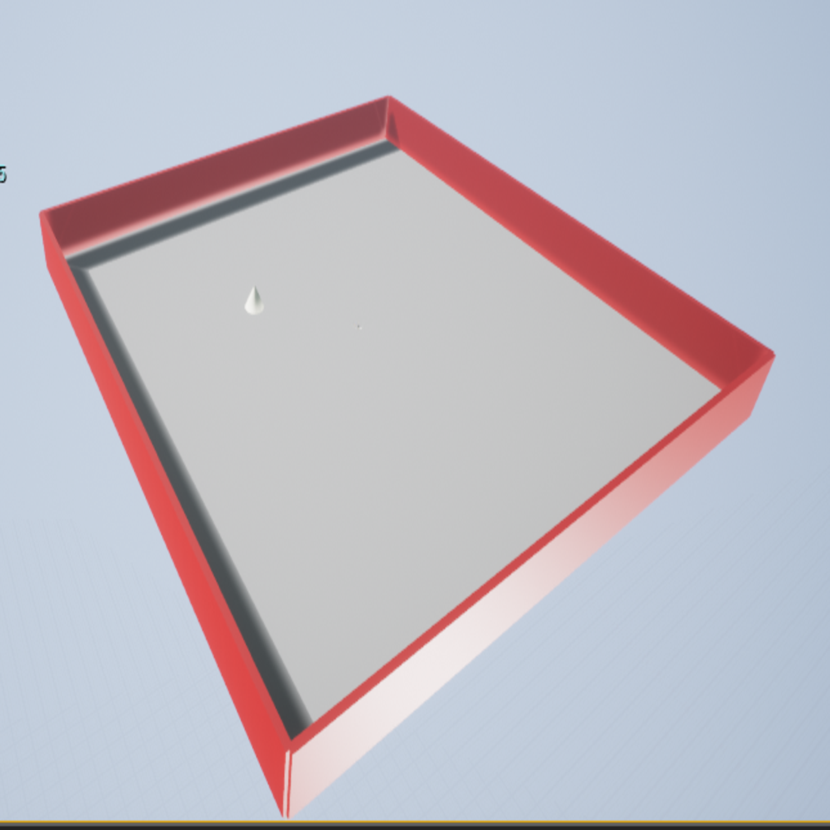}
        \caption{Arena with crimson walls.}
        \label{fig:arena}
    \end{subfigure}
    \hspace{10pt}
    \begin{subfigure}{0.3\linewidth}
        \includegraphics[width=\textwidth]{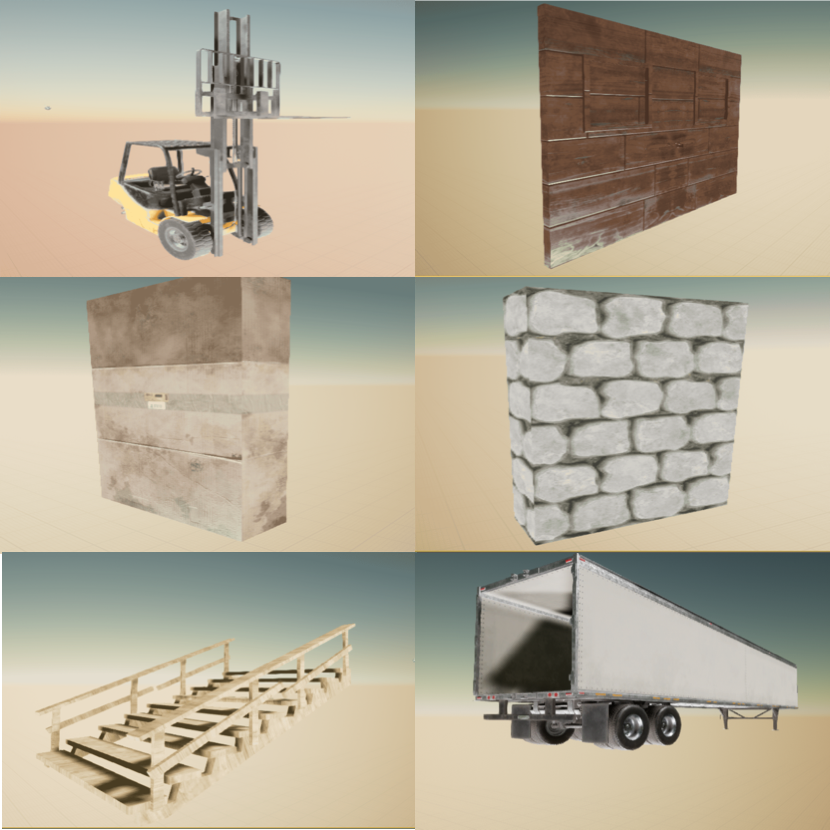}
        \caption{Obstacles.}
        \label{fig:assets}
    \end{subfigure}
    \hspace{10pt}
    \begin{subfigure}{0.3\linewidth}
        \includegraphics[width=\textwidth]{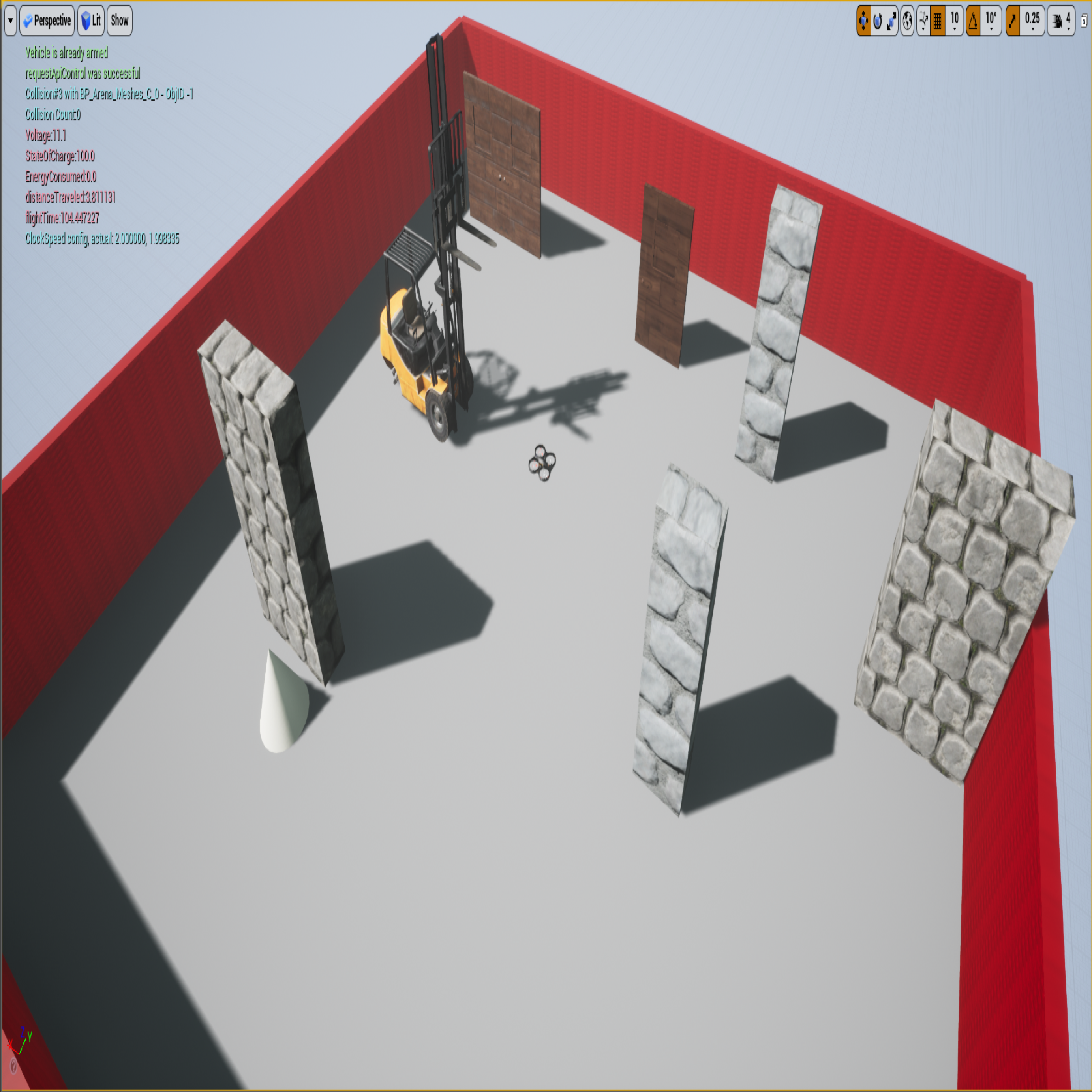}
        \caption{Arena with obstacles.}
        \label{fig:arena-obstacles}
    \end{subfigure}
    \caption{The environment generator generates different arena sizes with configurable wall texture colors, obstacles, obstacle materials etc. (a) arena with crimson colored walls with dimensions \texttt{50~m X 50~m X 5~m}. The arena can be small or several miles long. The wall texture color is specified as an [R, G, B] tuple, which allows the generator to create any color in the visible spectrum. (b) some of the UE4 asset used in \airl. Any UE4 asset can be imported and \airl environment generator will randomly select and spawn it in the arena. (c) arena with random obstacles. The positions of the obstacles can be changed every episode or a rate specified by the user.}
\end{figure*}

\subsection{Algorithm Exploration}
\label{sec:algo-exp}

Deep reinforcement learning is still a nascent field that is rapidly evolving. Hence, there is significant infrastructure overhead to integrate random environment generator and evaluate new deep reinforcement learning algorithms for UAVs.

So, we expose our random environment generator and AirSim UE4 plugin as an OpenAI gym interface and integrate it popular reinforcement learning framework with stable baselines~\cite{stable-baselines}, which is based on OpenAI baselines.\footnote{We also support Keras-RL, another widely used RL framework.} To expose our random environment generator into an OpenAI gym interface, we extend the work of AirGym~\cite{airgym-github} to add support for environment randomization, a wide range of sensors (Depth image, Inertial Measurement Unit (IMU) data, RGB image, etc.) from AirSim and support exploring multimodal policies.

We seed the \airl algorithm suite with two popular and commonly used reinforcement learning algorithms. The first is Deep Q Network (DQN)~\cite{dqn} and the second is Proximal Policy Optimization (PPO)~\cite{ppo-paper}. DQN falls into the discrete action algorithms where the action space is high-level commands (`move forward,' `move left' e.t.c.,) and Proximal Policy Optimization falls into the continuous action algorithms (e.g., policy predicts the continuous value of velocity vector). For each of the algorithm variants, we also support an option to train the agent using curriculum learning~\cite{bengio2009curriculum}. For both these algorithms, we keep the observation space, policy architecture and reward structure same and compare agent performance.
\blue{The environment configuration used in the training of PPO/DQN, the policy architecture, the reward function, is described in the appendix (Appendix~B).}

\blue{\Fig{fig:ppo-dqn-nc} shows the normalized reward of the DQN agent (DQN-NC) and PPO agent (PPO-NC) trained using non-curriculum learning. One of the observations is that the PPO agent trained using non-curriculum learning consistently accrues negative reward throughout the training duration. In contrast, the DQN agent trained using non-curriculum learning starts at the same as the PPO agent but the DQN agent accrues more reward beginning in the 2000\textsuperscript{th} episode.}

\blue{\Fig{fig:ppo-dqn-c} shows the normalized episodic reward for the DQN (DQN-C) and PPO (PPO-C) agents trained using curriculum learning. We observe a similar trend as we saw with the agents trained using non-curriculum learning where the DQN agent outperforms the PPO agent. However, in this case, the PPO agent has a positive total reward. But the DQN agent starts to accrue more reward starting from the 1000\textsuperscript{th} episode. Also, the slight dip in the reward at 3800\textsuperscript{th} is due to the curriculum's change (increased difficulty).} 

\begin{figure}[t!]
\centering
\begin{subfigure}{0.49\linewidth}
        \includegraphics[width=0.98\columnwidth,keepaspectratio]{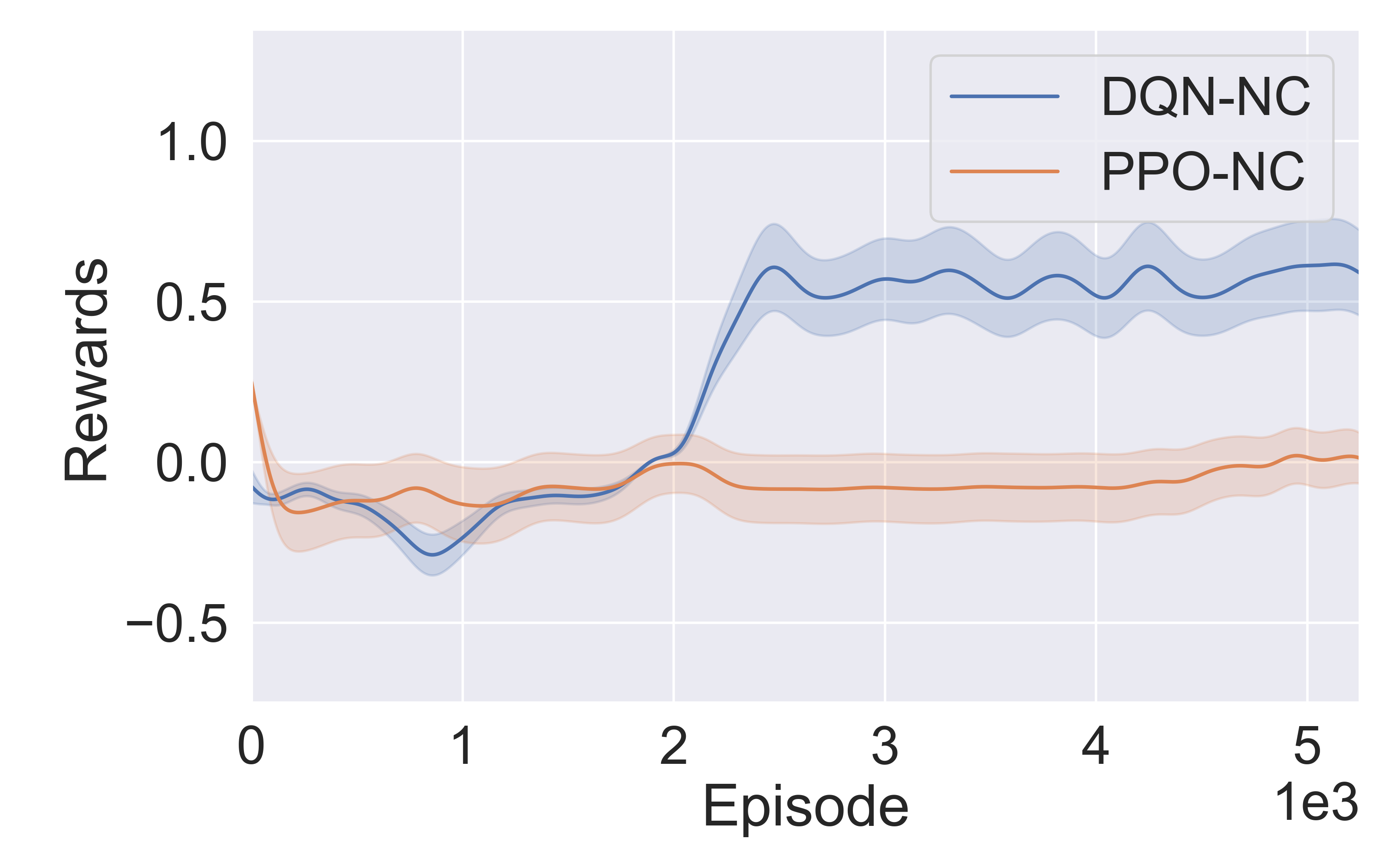}
        \caption{Non-curriculum learning.}
        \label{fig:ppo-dqn-nc}
        \end{subfigure}
        \begin{subfigure}{0.49\linewidth}
        \includegraphics[width=0.98\columnwidth, keepaspectratio]{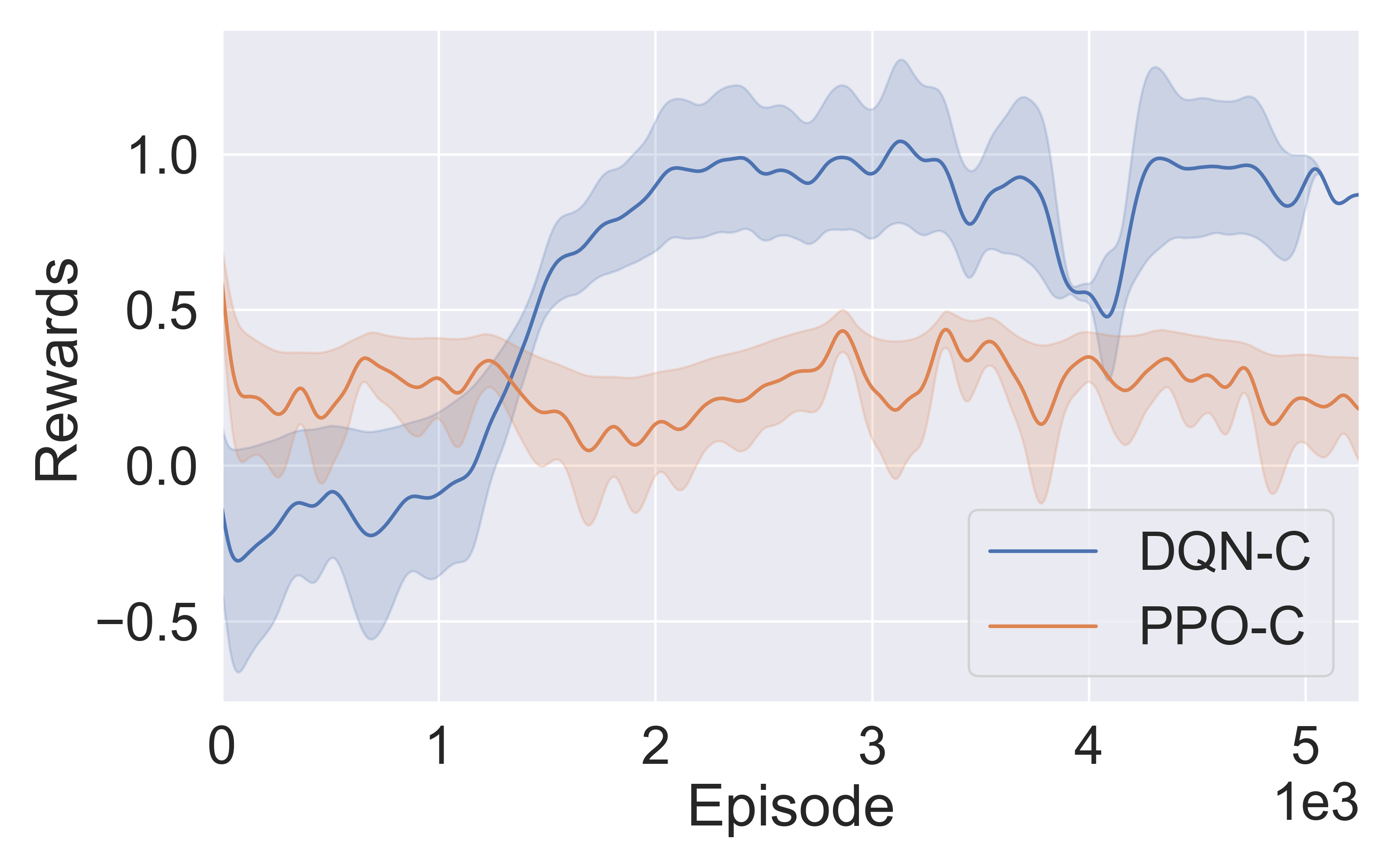}
        \caption{Curriculum learning.}
        \label{fig:ppo-dqn-c}
        \end{subfigure}
        \caption{(a) Normalized reward during training for algorithm exploration between PPO-NC and DQN-NC. (b) Normalized reward during training for algorithm exploration between PPO-C and DQN-C. We find that the DQN agent performs better than the PPO agent irrespective of whether the agent was trained using curriculum learning or non-curriculum learning.  \blue{The rewards are averaged over five runs with random seeds.}}

    \label{fig:ppo_vs_dqn}
\end{figure}

\blue{Reflecting on the results, we gathered in \Fig{fig:ppo-dqn-nc} and \Fig{fig:ppo-dqn-c}, continuous action reinforcement learning algorithms such as PPO have generally been known to show promising results for low-level flight controller tasks that are used for stabilizing UAVs~\cite{rl-uav-control}. However, as our results indicate, applying these algorithms for a complex task, such as end-to-end navigation in a photo-realistic simulator, can be challenging for a couple of reasons.} 

\blue{First, we believe that the action space for the PPO agent limits the exploration compared to the DQN agent. For the PPO agent, the action space is the components of velocity vector \texttt{v$_{x}$} and \texttt{v$_{y}$} whose value can vary from [-5$~m/s$, 5$~m/s$]. Having such an action space can be a constraining factor for PPO. For instance, if the agent observes an obstacle at the front, it needs to take action such that it moves right or left. Now for PPO agent, since the action space is continuous values of [\texttt{V$_{x}$}, \texttt{V$_{y}$}], for it to move forward in the $x$-direction, the \texttt{V$_{x}$} can be any positive number while the \texttt{V$_{y}$} component has to be `0'. It can be quite challenging for the PPO agent (or continuous action algorithm) to learn this behavior, and it might require a much more sophisticated reward function that identifies these scenarios and rewards or penalizes these behaviors accordingly. In contrast, for the DQN agent, the action space is much simpler since it has to only yaw (i.e., move left or right) and then move forward or vice versa.}

\blue{Second, in our evaluation, we keep the reward function, input observation and the policy architecture same for DQN and PPO agent. We choose to fix these because we want to focus on showcasing the capability of the \airl infrastructure. Since RL algorithms are sensitive to hyperparameters and the choice of the reward function, it could be possible that our reward function, policy architecture could have inadvertently favored the DQN agent compared to the PPO agent. The sensitivity of the RL algorithms to the policy and reward is still an open research problem~\cite{reward-shaping,reward-shaping-il}.}

\blue{The takeaway is that we can do algorithm exploratory studies with \airl. For high-level task like point-to-point navigation, discrete action reinforcement learning algorithms like DQN allows more flexibility compared to continuous action reinforcement learning algorithms like PPO. We also demonstrate that incorporating techniques such as curriculum learning can be beneficial to the overall learning.}

\subsection{Policy Exploration}
Another essential aspect of deep reinforcement learning is the policy, which determines the best action to take. Given a particular state the policy needs to maximize the reward. A neural network approximates the policies. To assist the researchers in exploring effective policies, we use Keras/TensorFlow~\cite{keras} as the machine learning back-end tool. Later on, we demonstrate how one can do algorithm and policy explorations for tasks like autonomous navigation though \airl is by no means limited to this task alone.

\subsection{Hardware Exploration}
\label{sec:hw-exp}

Often aerial roboticists port the algorithm onto UAVs to validate the functionality of the algorithms. These UAVs can be custom built~\cite{nvidia-ai-iot} or commercially available off-the-shelf (COTS) UAVs~\cite{ascending,intel} but mostly have fixed hardware that can be used as onboard compute. A critical shortcoming of this approach is that the roboticist cannot experiment with hardware changes. More powerful hardware may (or may not) unlock additional capabilities during flight, but there is no way to know until the hardware is available on a real UAV so that the roboticist can physically experiment with the platform. 

Reasons for wanting to do such exploration includes understanding the computational requirements of the system, quantifying the energy consumption implications as a result of interactions between the algorithm and the hardware, and so forth. Such evaluation is crucial to determine whether an algorithm is, in fact, feasible when ported to a real UAV with a specific hardware configuration and battery constraints.

For instance, a Parrot Bepop~\cite{parrot-bepop} comes with a P7 dual-core CPU Cortex A9 and a Quad core GPU. It is not possible to fly the UAV assuming a different piece of hardware, such as the NVIDIA Xavier~\cite{nvidia-xavier} processor that is significantly more powerful; at the time of this writing there is no COTS UAV that contains the Xavier platform. So, one would have to wait until a commercially viable platform is available. However, using \airl, one can experiment how the UAV would behave with a Xavier since the UAV is flying virtually. 

Hardware exploration in \airl allows for evaluation of the best reinforcement learning algorithm and its policy on different hardware. It is not limited by the onboard compute available on the real robot. Once the best algorithm and policy are determined, \airl allows for characterizing the performance of these algorithms and policies on different types of hardware platforms. It also enables to carefully fine-tune and co-design algorithms and policy while being mindful of the resource constraints and other limitation of the hardware.

\begin{figure}[t!]
\centering
        \includegraphics[width=0.95\columnwidth,keepaspectratio]{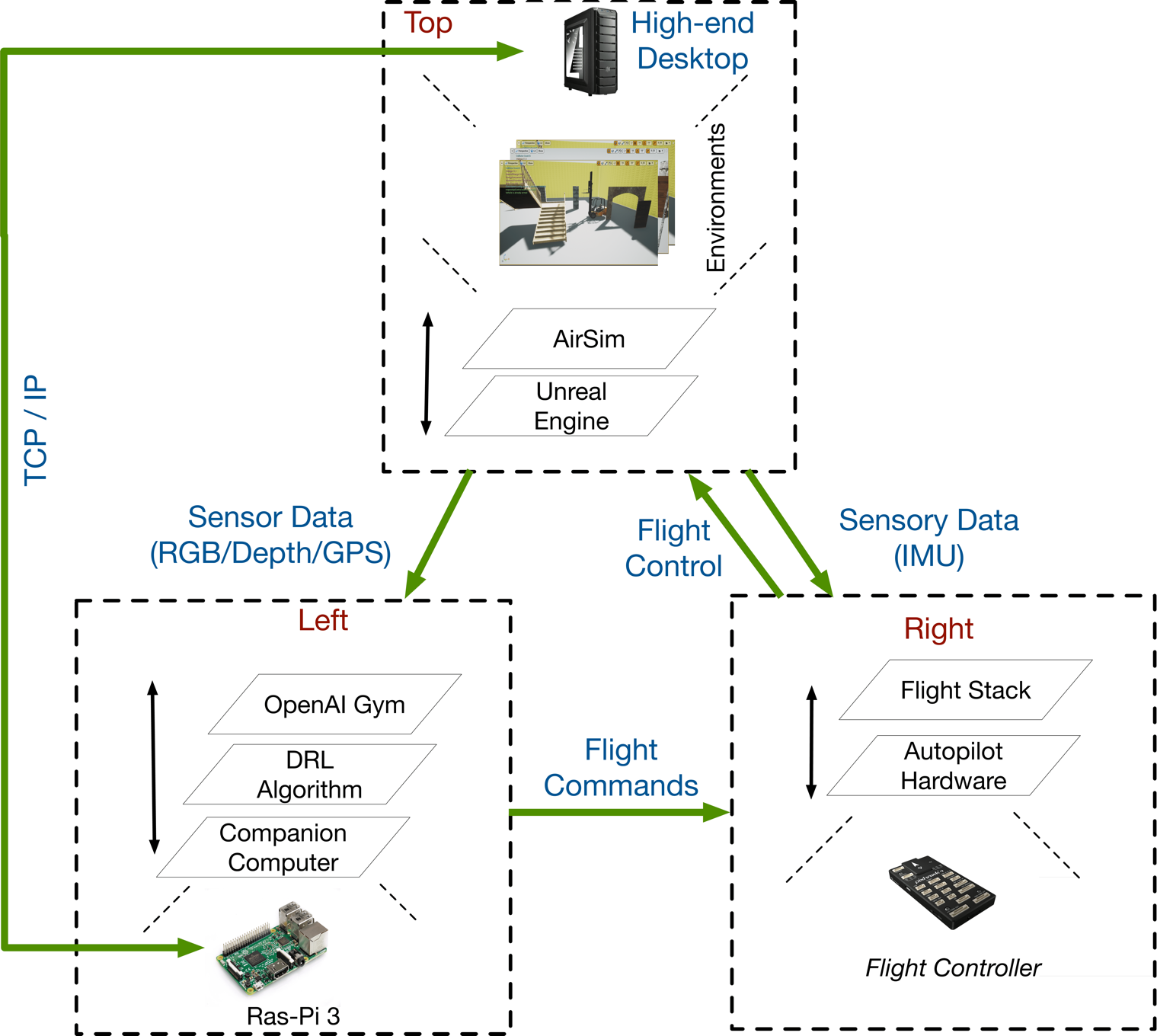}
        \caption{\small Hardware-in-the-loop (HIL) simulation in \airl.}

    \label{fig:hil}
\end{figure}

A HIL simulation combines the benefits of the real design and the simulation by allowing them to interact with one another as shown in \Fig{fig:hil}. There are three core components in \airl's HIL methodology: (1) a high-end desktop that simulates a virtual environment flying the UAV ($ top $); (2) an embedded system that runs the operating system, the deep reinforcement learning algorithms, policies and associated software stack ($ left $); and (3) a flight controller that controls the flight of the UAV in the simulated environment ($ right $).

The simulated environment models the various sensors (RGB/Depth Cameras), actuators (rotors), and the physical world surrounding the agent (Obstacles). This data is fed into the reinforcement learning algorithms that are running on the embedded companion computer, which processes the input and outputs flight commands to the flight controller. The controller then communicates those commands into the virtual UAV flying inside the simulated game environment.

The interaction between the three components is what allows us to evaluate the algorithms and policy on various embedded computing platforms. The HIL setup we present allows for the swap-ability of the embedded platform under test. The methodology enables us to effectively measure both the performance and energy of the agent holistically and more accurately, since one can evaluate how well an algorithm performs on a variety of different platforms. 

In our evaluation, which we discuss later, we use a Raspberry~Pi (\ras) as the embedded hardware platform to evaluate the best performing deep reinforcement learning algorithm and its associated policy. The HIL setup includes running the environment generator on a high-end desktop with a GPU. The reinforcement learning algorithm and its associated policy run on the \ras. The state information (Depth image, RGB image, IMU) are requested by Ras-Pi~3 using AirSim Plugins APIs which involves an RPC (remote procedural calls) over TCP/IP network (both high-end desktop and \ras are connected by ethernet). The policy evaluates the actions based on the state information it received from the high-end desktop. The actions are relayed back to the high-end desktop through AirSim flight controller API's. 

 \subsection{Energy Model in AirSim Plugin}In \airl, we use the energy simulator we developed in our prior work~\cite{mav-bench}. The AirSim plugin is extended with a battery and energy model. The energy model is a function of UAVs velocity, acceleration. The values of velocity and acceleration are continuously sampled and using these we estimate the power as proposed in this work~\cite{power-model}. The power is calculated using the following formula:
\newcommand{\norm}[1]{\left\lVert#1\right\rVert} 
\begin{equation}
\begin{aligned}
P = \begin{bmatrix}
		\beta_{1} \\
		\beta_{2} \\
		\beta_{3}
	\end{bmatrix}^{T}
    \begin{bmatrix}
		\norm{\vec{v}_{xy}} \\
		\norm{\vec{a}_{xy}} \\
		\norm{\vec{v}_{xy}}\norm{\vec{a}_{xy}}
	\end{bmatrix}
    +
    \begin{bmatrix}
		\beta_{4} \\
		\beta_{5} \\
		\beta_{6}
	\end{bmatrix}^{T}
    \begin{bmatrix}
		\norm{\vec{v}_{z}} \\
		\norm{\vec{a}_{z}} \\
		\norm{\vec{v}_{z}}\norm{\vec{a}_{z}}
	\end{bmatrix}
    \\
    +
    \begin{bmatrix}
		\beta_{7} \\
		\beta_{8} \\
		\beta_{9}
	\end{bmatrix}^{T}
    \begin{bmatrix}
		m \\
		\vec{v}_{xy} \cdot \vec{w}_{xy} \\
		1
	\end{bmatrix}
\end{aligned}
\label{eqn:power}
\end{equation}

In Eq.~\ref{eqn:power}, {v}$_{xy}$ and {a}$_{xy}$ are the velocity and acceleration in the horizontal direction. {v}$_{z}$ and {a}$_{z}$ denotes the velocity and acceleration in the $z$ direction. $m$ denotes the mass of the payload. $\beta_1$ to $\beta_9$ are the coefficients based on the model of the UAV used in the simulation. For the energy calculation model, we use the columb counter technique as described in prior work~\cite{columb-count}. The simulator computes the total number of columb that has passed over the battery over every cycle.

Using the energy model \airl allows us to monitor the energy continuously during training or during the evaluation of the reinforcement learning algorithm.

\subsection{Quality of Flight Metrics}
\label{sec:qof}

Reinforcement learning algorithms are often evaluated based on success rate where the success rate is based on whether the algorithm completed the mission. This metric only captures the functionality of the algorithm and grossly ignores how well the algorithm performs in the real world. In the real world, there are additional constraints for a UAV, such as the limited onboard compute capability and battery capacity.

Hence, we need additional metrics that can quantify the performance of learning algorithms more holistically. To this end, \airl introduces Quality-of-Flight (QoF) metrics that not only captures the functionality of the algorithm but also how well they perform when ported to onboard compute in real UAVs. For instance, the algorithm and policies are only useful if they accomplish the goals within finite energy available in the UAVs.
Hence, algorithms and policies need to be evaluated on the metrics that describe the quality of flight such as mission time, distance flown, etc. In the first version of \airl, we consider the following metrics.

\textbf{Success Rate:} The percentage of time the UAV reaches the goal state without collisions and running out of battery. Ideally, this number will be close to 100\% as it reflects the algorithms' functionality, taking into account resource constraints.

\textbf{Time to Completion:} The total time UAV spends finishing a mission within the simulated world.

\textbf{Energy Consumed:} The total energy spent while carrying out the mission. Limited battery available onboard constrains the mission time. Hence, monitoring energy usage is of utmost importance for autonomous aerial vehicles, and therefore should be a measure of policy's efficiency.

\textbf{Distance Traveled:} Total distance flown while carrying out the mission. This metric is the average length of the trajectory that can be used to measure how well the policy did.

\subsection{Runtime System}

The final part is the runtime system that orchestrates the overall execution. The runtime system starts the game engine with the correct configuration of the environment before the agent starts. It also monitors the episodic progress of the reinforcement learning algorithm and ensures that before starting a new episode that it randomizes the different parameters, so the agent statistically gets a new environment. It also has resiliency built into it to resume the training in case any one of the components (for example UE4 engine) crashes.

In summary, using \airl environment generator, researchers can develop various challenging scenarios to design better learning algorithms. Using \airl interfaces to OpenAI gym, stable-baselines and TensorFlow backend, they can rapidly evaluate different reinforcement learning algorithms and their associated policies. Using \airl HIL methodology and QoF metrics, they can benchmark the performance of learning algorithms and policies on resource-constrained onboard compute platforms. 

%% file: prelude.tex
\section{Experimental Evaluation Prelude}

The next few sections focus heavily on how \airl can be used to demonstrate its value. As a prelude, this section presents the highlights to focus on the big picture.

\emph{Policy Evaluation (Section~\ref{sec:policy-exploration}):} We show how \airl can be used to explore different reinforcement learning based policies. We use the best algorithm determined during the algorithm exploration step and use that algorithm to explore the best policy. In this work, we use \airl environment generator to generate three environments namely \noobj, \staticobj, and \dynamicobj.  These three environments create a varying level of difficulty by changing the number of static and dynamic obstacles in the environments for the autonomous navigation task. 
 \begin{figure}[t!]
    \begin{subfigure}{.42\linewidth}
        \includegraphics[width=\textwidth]{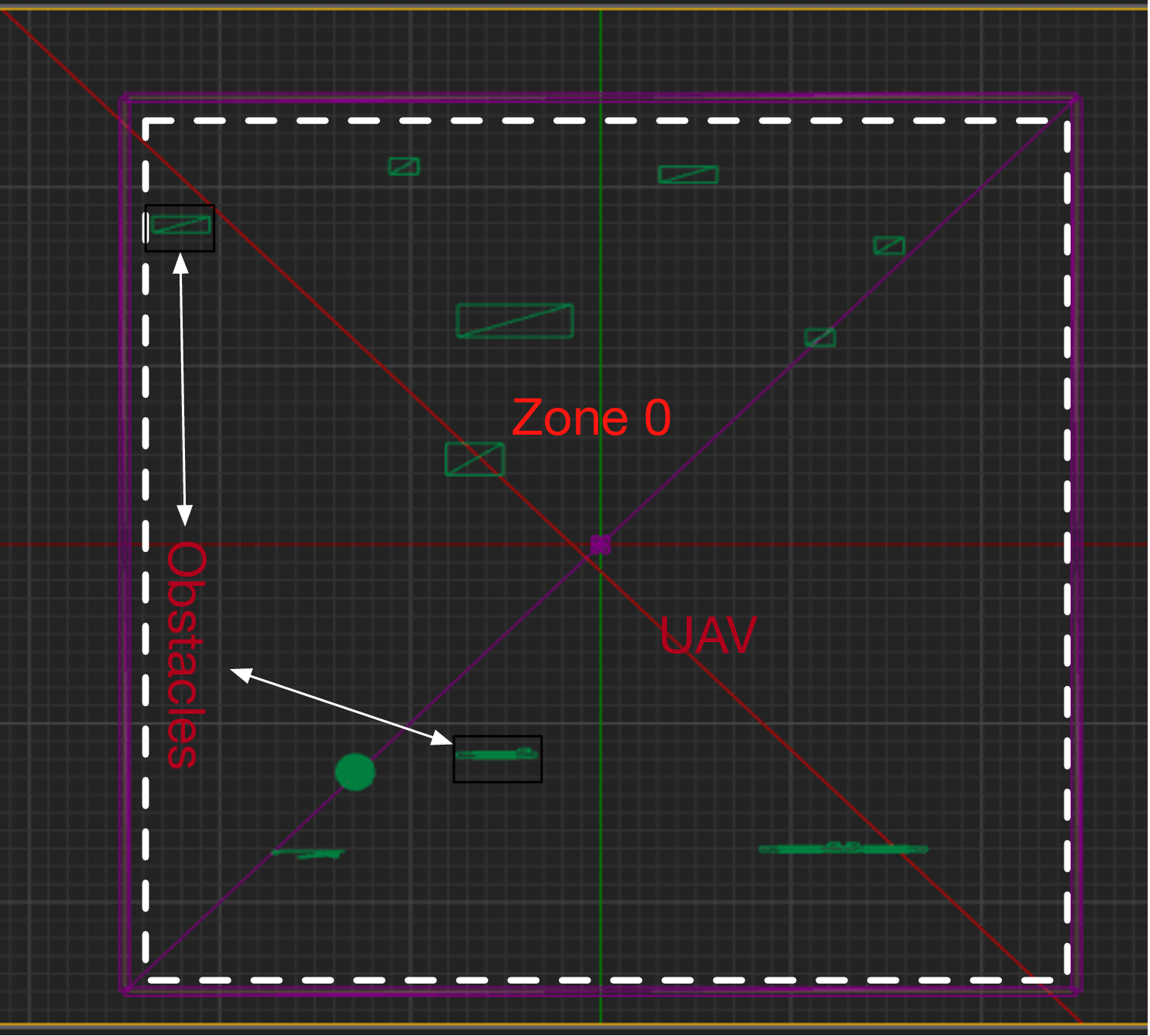}
        \caption{\footnotesize Non-curriculum learning based on static obstacles.}
        \label{fig:ncr_zone}
    \end{subfigure}
    \hspace{10pt}
    \begin{subfigure}{.42\linewidth}
        \includegraphics[width=\textwidth]{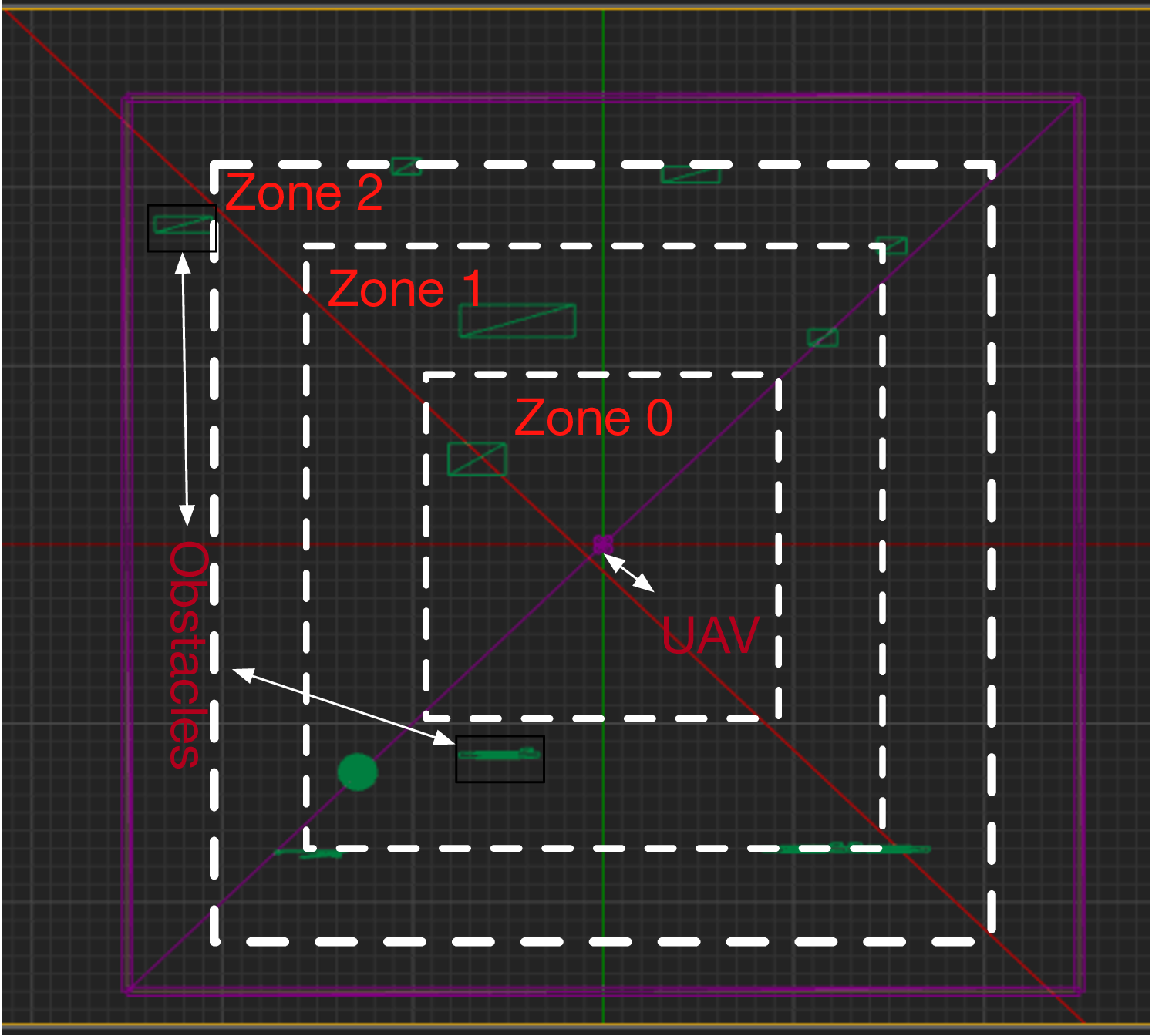}
        \caption{\footnotesize Curriculum learning based on static obstacles.}
        \label{fig:cur_zone}
    \end{subfigure}
    \caption{Zoning used in the training methodology for curriculum learning and non-curriculum learning. Here we show the top view of our environment in wireframe mode~\cite{wire-frame} available in UE4. (a) In non-curriculum learning, the end goal is randomly placed anywhere in the arena. Unlike curriculum learning, the entire arena is one zone. (b) In curriculum learning, we split the arena into virtual partitions, and the end goal is placed within a specific zone and gradually moved higher zone once it succeeds in more than 50\% over 1000 latest episode. }
 \end{figure}
We also show how \airl allows end users to perform benchmarking of the policies by showing two examples. In the first example, we show how well the policies trained in one environment generalize to the other environments. In the second example, we show to which of the sensor inputs the policy is most sensitive towards. This insight can be used while designing the network architecture of the policy. For instance, we show that image input has the highest sensitivity amongst other inputs. Hence a future iteration of the policy can have more feature extractors (increasing the depth of filters) dedicated to the image input.

\emph{System Evaluation (Section~\ref{sec:sys-eval}):} We show the importance of benchmarking algorithm performance on resource-constrained hardware such as what is typical of a UAV compute platform. In this work, we use a Raspberry~Pi~4 (\ras) as an example of resource-constrained hardware. We use the best policies determined in the policy exploration step (Section~\ref{sec:policy-exploration}) and use that to compare the performance between Intel Core-i9 and \ras using HIL and the QoF metrics available in \airl. We also show how to artificially degrade the performance of the Intel Core-i9 to show how compute performance can potentially affect the behavior of a policy when it is ported over to a real aerial robot.

In summary, using these focused studies, we demonstrate how \airl can be used by researchers to design and benchmark algorithm-hardware interactions in autonomous aerial vehicles, as shown previously in \Fig{fig:airlearning-infrastructure}.

%% file: policy_exploration.tex
\section{Policy Exploration}
\label{sec:policy-exploration} 

In this section, we perform policy exploration for the DQN agent with curriculum learning~\cite{bengio2009curriculum}. The policy exploration phase aims to determine the best neural network policy architecture for each of the tasks (i.e., autonomous navigation) in different environments with and without obstacles. 

\begin{figure*}[t!]
\centering
        \includegraphics[width=0.9\columnwidth,keepaspectratio]{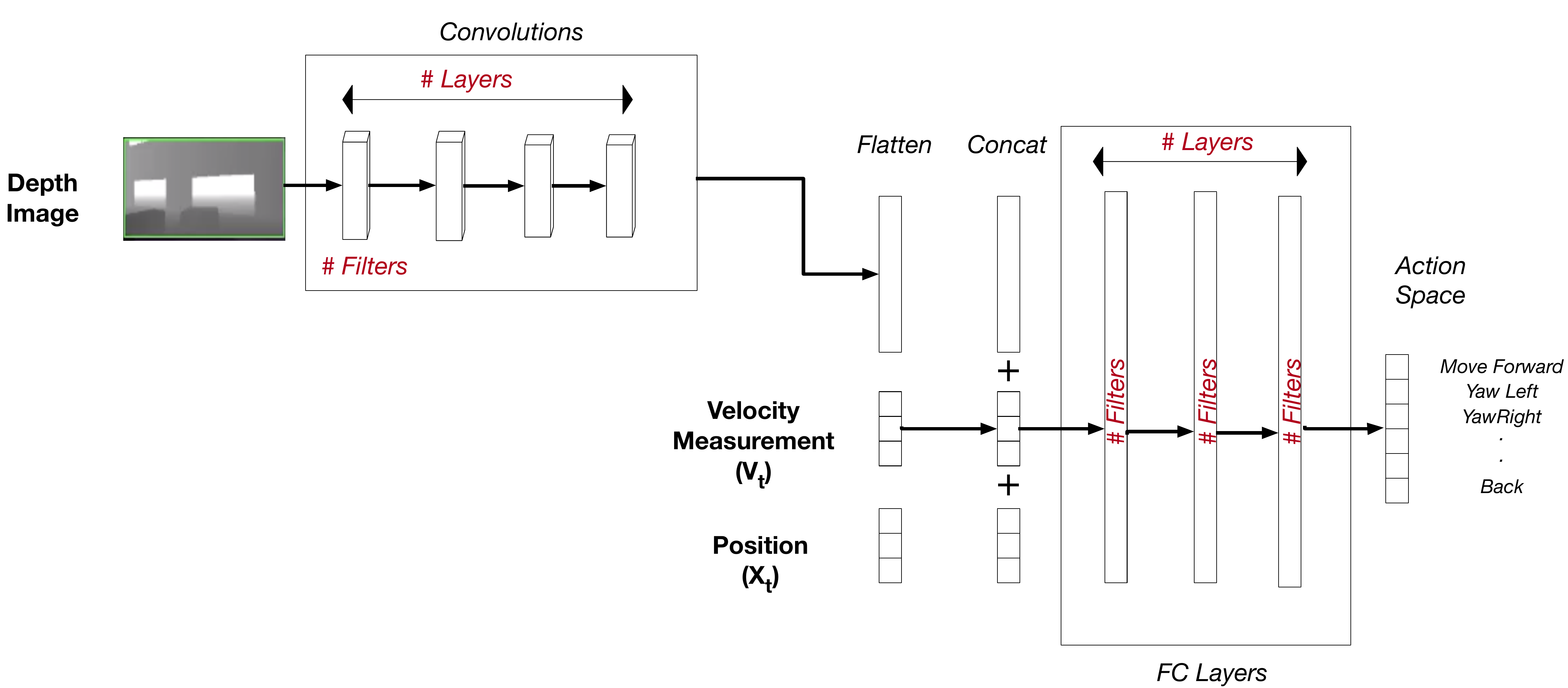}
        \caption{The network architecture template for the policies used in DQN agents. We sweep the \textit{\# Layers} and \textit{\# Filters} parameters in the network architecture template. Both the agents take a depth image, velocity vector, and position vector as inputs. The depth image is passed through \# Layers of convolutions layers with \# Filters each. \# Layers and \# Filters are variables what we sweep. We also use a uniform filter size of \texttt{(3 x 3)} with stride of 1. 
        The combined vector space is passed to the \textit{\# Layers} of fully connected network, each with \textit{\# Filters} hidden units. \blue{The choice of using \textit{\#Layers} and \textit{\# Filters} parameters to control both the convolution and fully-connected layers is to manage the complexity of searching over large NN hyperparameters design space.} The action space determines the number of hidden units in the last fully connected layer. For the DQN agent, we have twenty-five actions.}

    \label{fig:policy-arch-search}
\end{figure*}

We start with a basic template architecture, as shown in \Fig{fig:policy-arch-search}. The  architecture is multi-modal and takes depth image, velocity, and position data as its input. Using this template, we sweep two parameters, namely \textit{\# Layers and \# Filters} (making the policy wider and deeper). To simplify the search, for convolution layers, we restrict filter sizes to 3 x 3 with stride 1. This choice ensures that there is no loss of pixel information. Likewise, for fully-connected layers, \# Filter parameter denotes the number of hidden neurons in that layer. \blue{The choice of using \textit{\#Layers} and \textit{\# Filters} parameters to control both the convolution and fully-connected layers is to manage the complexity of searching over large NN hyperparameters design space.}

The \textit{\# Layers and \# Filters} and the template policy architecture can be used to construct a variety of different policies. For example, a tuple of (\# Filters $=$ 32, \# Layers $=$ 5) will result in a policy architecture where there five convolution layers with 32 filters (with 3 x 3 filters) followed by five fully-connected layers with 32 hidden neurons each. For each of the navigation tasks (in different environments), we sweep the template parameters (\# Layers and \# Filters) to explore multiple policy architectures for the DQN agent.

\subsection{Training and Testing Methodology}
The training and testing methodology for the DQN  agent running in the different environments is described below. 

\textbf{Environments:} For the point-to-point autonomous navigation task for UAVs, we create three randomly generated environments, namely \noobj, \staticobj, and \dynamicobj with varying levels of static obstacles and dynamic obstacles. The environment size for all three levels is 50~m x 50~m. For the \noobj environment, there are no obstacles in the main arena, but the goal position is changed every episode. For \staticobj, the number of obstacles varies from five to ten, and it is changed every four~episodes. The end goal and position of the obstacles are changed every episode. For \dynamicobj, along with five static obstacles, we introduce up to five dynamic obstacles of whose velocities range from 1~$m/s$ to 2.5~$m/s$. The obstacles and goals are placed in random locations every episode to ensure that the policy does not over-fit.

\textbf{Training Methodology:} We train the DQN agent using curriculum learning in the environments described above. We use the same methodology described in Appendix~B, where we checkpoint policy in each zone for the three environments. The hardware used in training is an Intel Core-i9 CPU with an Nvidia GTX 2080-TI GPU.

\textbf{Testing Methodology:} 
For testing the policies, we evaluate the checkpoints saved in the final zone. \blue{Each policy is evaluated on 100 randomly generated goal/obstacle configuration (controlled by the \textit{`Seed'} parameter in Table~\ref{tab:game-config}). The same 100 randomly generated environment configurations are used across different policy evaluations.} The hardware we use for testing the policies is the same as the hardware used for training them (Intel Core-i9 with Nvidia GTX 2080-TI).

\begin{figure*}[t!]
    \begin{subfigure}{.33\linewidth}
        \includegraphics[width=\textwidth]{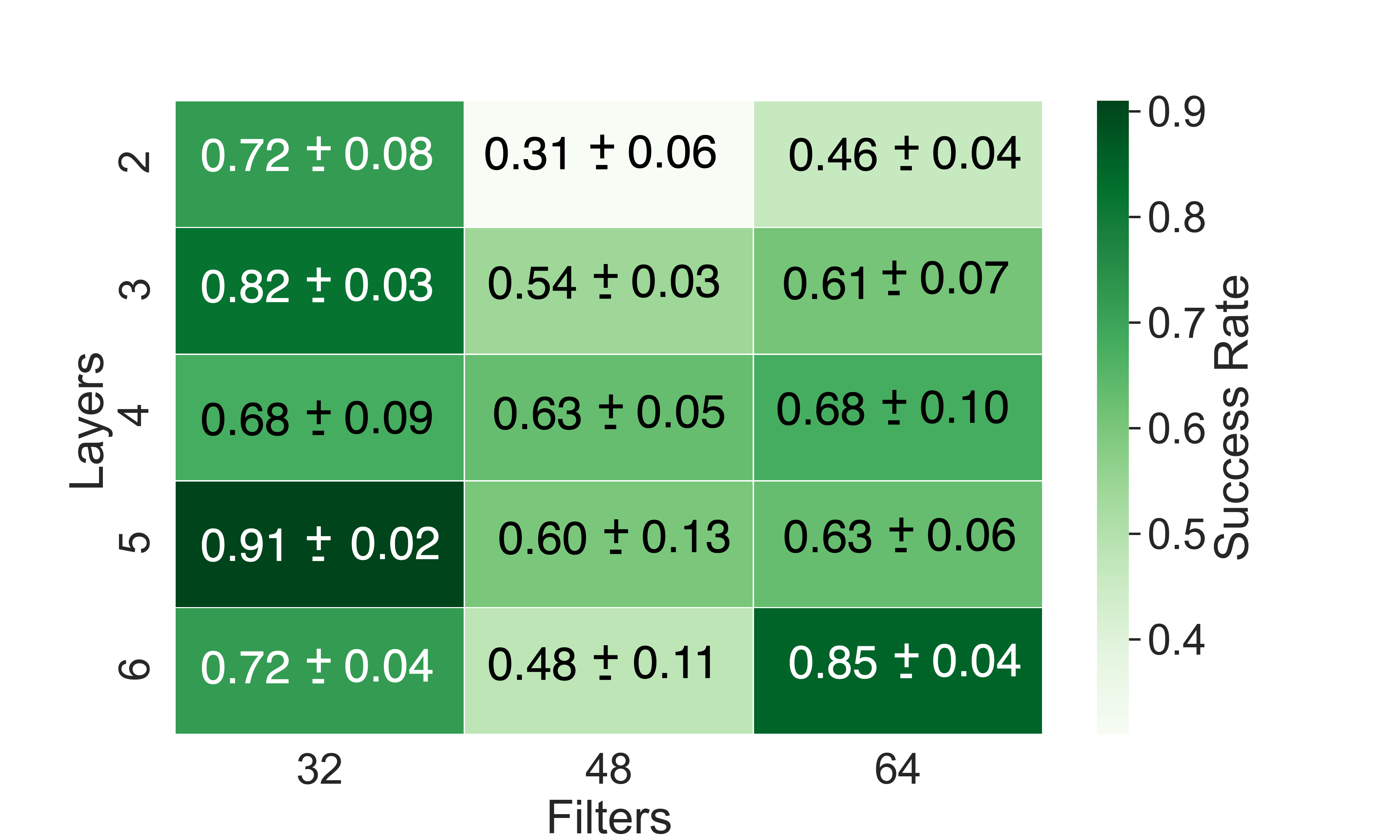}
        \caption{\footnotesize No obstacles.}
        \label{fig:cur_noobs}
    \end{subfigure}
    \hspace{-15pt}
    \begin{subfigure}{.33\linewidth}
        \includegraphics[width=\textwidth]{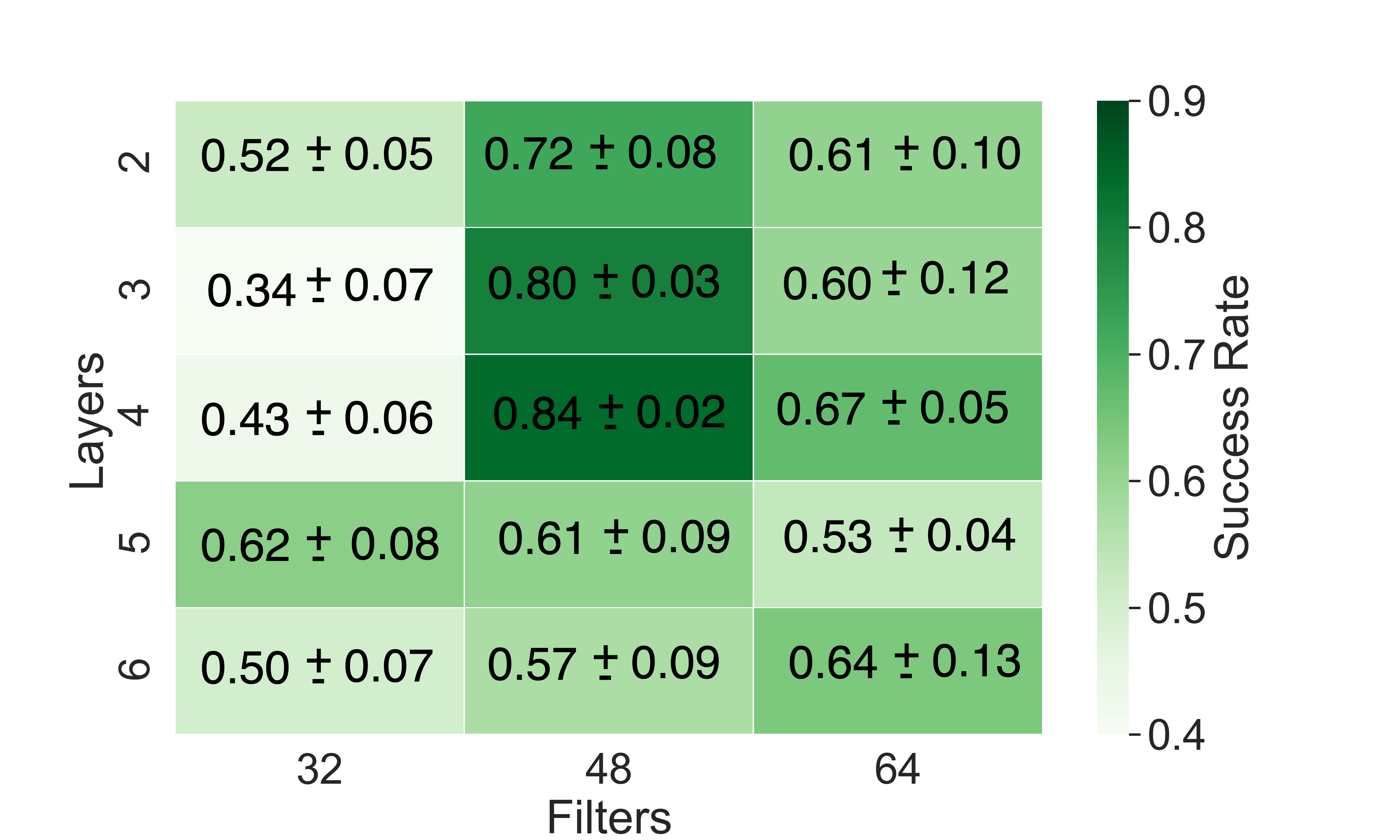}
        \caption{\footnotesize Static obstacles.}
        \label{fig:cur_statobs}
    \end{subfigure}
    \hspace{-15pt}
    \begin{subfigure}{.33\linewidth}
        \includegraphics[width=\textwidth]{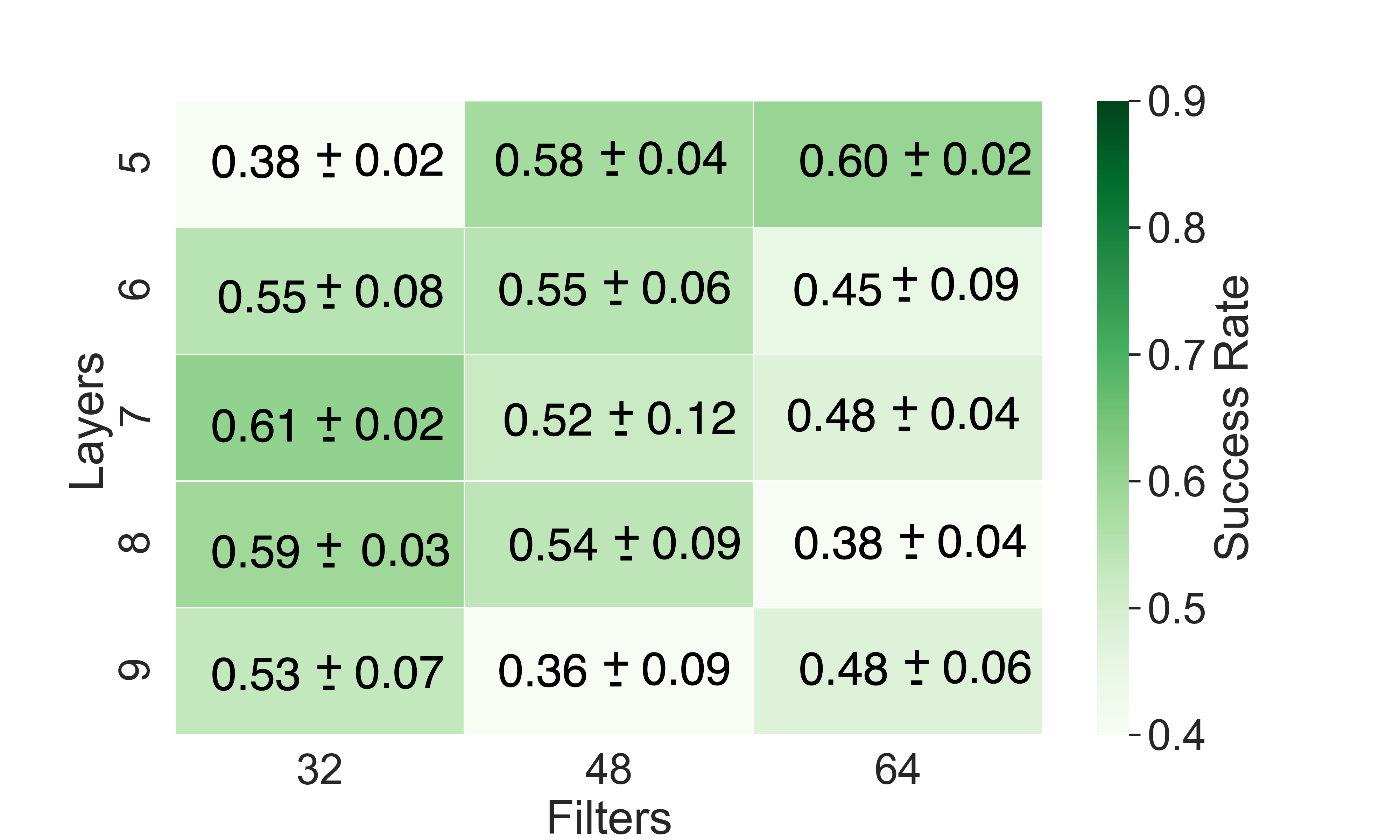}
        \caption{\footnotesize Dynamic obstacles.}
        \label{fig:cur_dynobs}
    \end{subfigure}
    \caption{(a), (b), and (c) show the policy architecture search for the \noobj, \staticobj, and \dynamicobj environments. Each \blue{cell} shows the success rate for the policies for \# Layers and \# Filters' corresponding values. The success rate is evaluated in \zoneD, which is the region that is not used during training. \blue{Each policy is evaluated on the same 100 randomly generated environment configuration (controlled by the \textit{`Seed'} parameter described in Table~\ref{tab:game-config}).} The policy architecture with the highest success rate is chosen as the best policy for DQN agents in the environment with no obstacles, static obstacles, and dynamic obstacles. \blue{The standard deviation error across multiple seeds are denoted by ($\pm$) sign}. For the \noobj environment, the policy with \# Layers of five and \# Filters of 32 is chosen as the best performing policy. Likewise, for the \dynamicobj environment, the policy architecture with \# Layers of 7 and \# Filter of 32 is chosen as the best policy.}
    \label{fig:policy-eval}
   
\end{figure*}

%% file: policy-eval.tex
\subsection{Policy Selection}

The policy architecture search for \textit{No Obstacles}, \textit{Static Obstacles}, and \textit{Dynamic Obstacles} are shown in \Fig{fig:policy-eval}. \Fig{fig:cur_noobs}, \Fig{fig:cur_statobs} and \Fig{fig:cur_dynobs} show the success rate for different policy architecture searched for the DQN agent trained using curriculum learning on \noobj, \staticobj, and \dynamicobj environments, respectively. In the figures, the x-axis corresponds \# Filter sizes (32, 48, or 64) and the y-axis corresponds to the \# Layers (2, 3, 4, 5, and 6) \blue{for \noobj/\staticobj environments and \# Layers (5, 6, 7, 8, 9) for \dynamicobj environment. The reason for sweeping different (larger) policies is because ``Dynamic Obstacles'' will be a harder task, and a deeper policy might help improve the success rate compared to a shallow policy.} Each cell corresponds to a unique policy architecture based on the template defined in \Fig{fig:policy-eval}. \blue{The value in each cell corresponds to the success rate for the best policy architecture. The $\pm$ denotes the standard deviation (error bounds) across five seeds. For instance, in \Fig{fig:cur_noobs}, the best performing policy architecture with \# Filters of 32 and \# Layers of 2 results in a 72\% success rate. The success rate across five seeds results in a standard deviation of $\pm$ of 8\% error. For evaluation, we always choose the best performing policy (i.e., the policy that achieves best success rate).}

Based on the policy architecture search, we notice that as the task complexity increases (obstacle density increases), a larger policy improves the task success rate. For instance, in the \noobj case (\Fig{fig:cur_noobs}), the policy with \# Filters of 32 and \# Layers of 5 achieves the highest success rate of 91\%. Even though we name the environment No Obstacles, the UAV agent can still collide with the arena walls, which lowers the success rate. For \staticobj case (\Fig{fig:cur_statobs}), the policy with \# Filters of 48 and \# Layers of 4 achieves the best success rate of 84\%. Likewise, for \dynamicobj case (\Fig{fig:cur_dynobs}), the policy architecture with \# Filters of 32 and \# Layers of 7 achieves the best success rate of 61\%. The success rate loss in \staticobj and \dynamicobj cases can be attributed to an increase in the possibility of collisions with static and dynamic obstacles.

\subsection{Success Rate Across the Different Environments}

To study how a policy trained in one environment performs in other environments, we take the best policy trained in the \noobj environment and evaluate it on the \staticobj and \dynamicobj environments. We do the same for the best policy trained on \dynamicobj and assess it on the \noobj and \staticobj environments. 

The results for the generalization study are tabulated in Table~\ref{tab:generalization}. We see that the policy trained in the \noobj environment has a steep drop in success rate from 91\% to 53\% in \staticobj and 32\% in \dynamicobj environment, respectively. In contrast, we observe that the policy trained in the \dynamicobj environment has an increased success rate from 61\% to 89\% in the \noobj and 74\% in the \staticobj environment, respectively.
\begin{table}[t!]
\vspace{5pt}
\renewcommand*{\arraystretch}{2.0}
\resizebox{\columnwidth}{!}{%
\begin{tabular}{|l|r|r|r|}
\hline
\textbf{ \large Policy (\# Layers, \# Filters)}                        & \textbf{\large No Obstacles} & \textbf{\large Static Obstacles} & \textbf{\large Dynamic Obstacles} \\ \hline\hline
\textit{\large No Obstacles (5,32)}      & \large 0.91                        & \large 0.53                         & \large 0.32                          \\ \hline
\textit{\large Dynamic Obstacles (7,32)} & \large 0.89                     & \large 0.74                         & \large 0.61                          \\ \hline
\end{tabular}%
}
\caption{Evaluation of the best-performing policies trained in one environment tested in another environment. We evaluate the best performing policy (7 Layers, 32 Filters) trained on \dynamicobj in \noobj and \staticobj environment. Likewise, we also evaluate the best performing policy (5 Layers, 32 Filters) trained in the \noobj environment in the \staticobj and \dynamicobj environments.}
\label{tab:generalization}
\end{table}

The drop in the success rate for the policy trained in the \noobj environment is expected because, during its training, the agent might not have encountered a variety of obstacles (static and dynamic obstacles) to learn from as it might have encountered in the other two environments. The same reasoning can also apply to the improvement in the success rate observed in the policy trained in the \dynamicobj environment when it is evaluated on the \noobj and \staticobj environments.

In general, the agent performs best in the environment where it is trained, which is expected. But we also observe that training an agent in a more challenging environment can yield good results when evaluating in a much less challenging environment. Hence, having a random environment generator, such as what we have enabled in Air Learning, can help the policy generalize well by creating a wide variety of different experiences for the agent to experience during training.

\subsection{Success Rate Sensitivity to Sensor Input Ablation}

In doing policy exploration, one is also interested in studying the policy's sensitivity towards a particular sensor input. So we ablate the sensor inputs to the policy to understand the effects. We ablate the policy's inputs one by one and see the impact of various ablation and its success rate. \blue{It is important to note that we do not re-train the policy with ablated inputs. This is to perform reliability study and simulate the real-world scenario if a particular sensing modality is corrupted. }

\begin{figure}[t!]
\centering
\includegraphics[width=0.8\columnwidth,keepaspectratio]{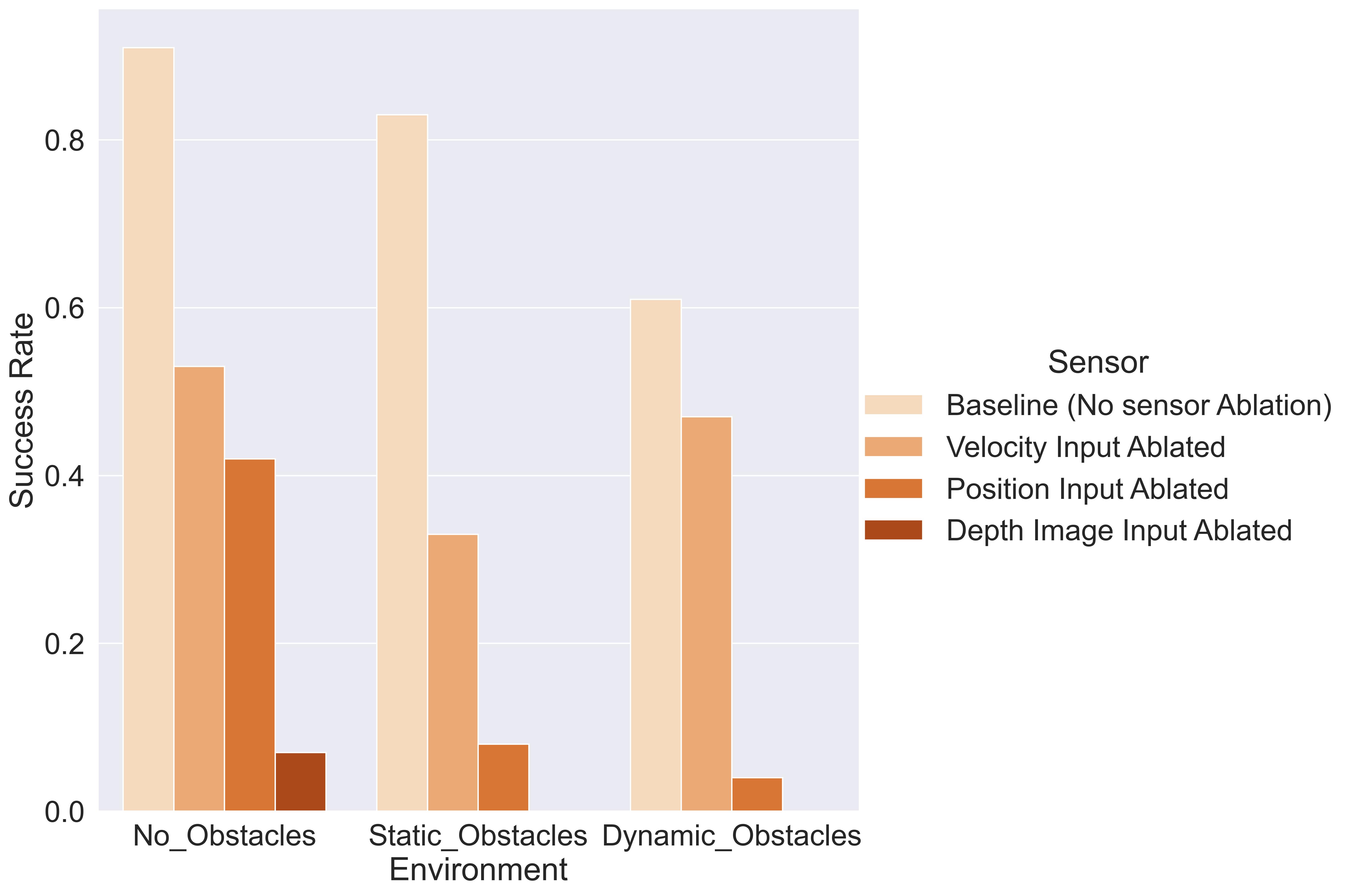}
\caption{The effect of ablating the sensor inputs on the success rate. We observe that the depth image contributes the most to the policy's success, whereas velocity input affects the least in the success. All the policy evaluations are in \texttt{Zone3} on the Intel Core-i9 platform.}
\label{fig:ablation-study}
\end{figure}

The policy architecture we used  for the DQN agent in this work is multi-modal in nature which receives depth image, velocity measurement \textit{V$_t$} and position vector \textit{X$_t$} as inputs. The \textit{V$_t$} is a 1-dimensional vector of the form [\textit{v$_x$}, \textit{v$_y$}, \textit{v$_z$}] where \textit{v$_x$}, \textit{v$_y$}, \textit{v$_z$} are the components of velocity vector in x, y and z directions at time \textit{`t'}. The \textit{X$_t$} is a 1-dimensional vector of the form [\textit{X$_{goal}$}, \textit{Y$_{goal}$}, \textit{D$_{goal}$}], where \blue{\textit{X$_{goal}$} and  \textit{Y$_{goal}$} are the relative `x' and `y' distance with respect to the goal position and \textit{D$_{goal}$} is the euclidean distance to the goal from the agent's current position.}   

The baseline success rate we use in this study is when all the three inputs are fed to the policy. The velocity ablation study refers to removing the velocity input measurements from policy inputs. Likewise, the position ablation study and depth image ablation study refer to removing the position vector and depth image from the policy's input stream. The results of various input ablation studies are plotted in \Fig{fig:ablation-study}.

For the \noobj environment, the policy success rate drops from 91\% to 53\% when velocity measurements are ablated. When the depth image is ablated, we find that the success rate drops to 7\%, and when the position vector is ablated, the success rate drops to 42\%. Similarly, for \staticobj, we find that if the depth image input is ablated, it fails to reach the destination. Likewise, when the velocity and position inputs are ablated, we observe the success rate drops from 84\% to 33\%. Similarly, we see a similar observation in a \dynamicobj environment where the success rate drops to 0\% when the depth image is ablated.

The depth image is the highest contributor to the policy's success, whereas the velocity input is significant but least among the other two inputs. \blue{The drop in the policy success rate due to depth image ablation is evident from policy architecture since maximum features in the flatten layer are contributed by the depth image than velocity and position (both 1 x 3 vectors). Another interesting observation is that when the position input is ablated, the agent also loses the information about its goal. The lack of goal position results in an exploration policy capable of avoiding obstacles (due to depth image input). In \noobj environment (where there are no obstacles except walls), the agent is free to explore unless it collides with the walls or exhaust maximum allowed steps. Due to the exploration, the agent reaches the goal position 42 out of 100 times. Our results are in line with prior work~\cite{source-seeking,random-actions} where such random action-based exploration yields some amount of success. However, in a cluttered environment, random exploration may result in sub-optimal performance due to a higher probability of collision or exhausting maximum allowed steps (a proxy for limited battery energy).}

Using Air Learning, researchers can gain better insights into how reliable a particular set of inputs in the case of sensor failures. The reliability studies and its impact on learning algorithms are essential given the kind of application the autonomous aerial vehicles are targeted. Also, understanding the sensitivity of a particular input towards success can lead to better policies where more feature extraction can be assigned to those inputs.

%% file: sys-eval.tex
\section{System Evaluation}

\label{sec:sys-eval}

This section demonstrates how Air Learning can benchmark the algorithm and policy's performance on a resource-constrained onboard compute platform, post-training. 
We use the HIL methodology (Section~\ref{sec:hw-exp}) and QoF metrics (Section~\ref{sec:qof}) for benchmarking the DQN agent and its policy. We evaluate them on the three different randomly generated environments described in Section~\ref{sec:policy-exploration}. 

\subsection{Experimental Setup}
\label{sec:sys-eval-setup}
\begin{figure}[t!]
        \includegraphics[width=0.9\textwidth]{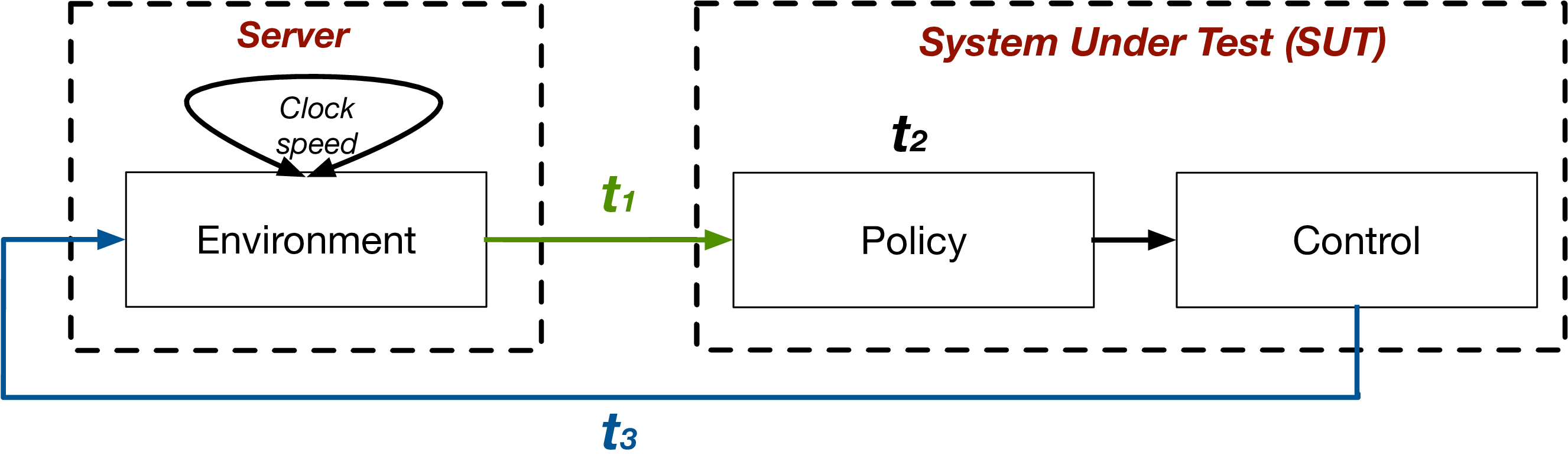}
        \caption{The Experimental setup for policy evaluation on two different platforms. The platform under test is called the System Under Test (SUT). The environments are rendered on a server with Intel Core-i9 with Nvidia RTX 2080. Clock speed is a function in the AirSim plugin, which speeds up the environment time relative to the real world clock. In our evaluation, we set the clock speed to 2X. Time t$_{1}$ is the time it takes to get the state information from the environment to the SUT. We use an Intel Core-i9 and a \ras as the two SUTs. Time t$_{2}$ is the time it takes to evaluate the forward pass of the neural network policy. This latency depends on the SUT. It is different for the Intel Core-i9 and the \ras. Time t$_{3}$ is the actuation time for which the control is applied. }
    \label{fig:hil-sys-eval}
 \end{figure}

The experimental setup has two components, namely, the server and System Under Test (SUT), as shown in \Fig{fig:hil-sys-eval}. The server component is responsible for rendering the environment (for example, \textit{\noobj}). The server consists of an 18 core Intel Core-i9 processor with an Nvidia RTX-2080. The SUT component is the system on which we want to evaluate the policy. The SUT is the proxy for the onboard compute system used in UAVs. In this work, we compare the policies' performance on two systems, namely Intel Core-i9 and \ras. The key differences between the Intel Core-i9 and \ras platform are tabulated in \Tab{tab:intel-ras-pi}. The systems are vastly different in their performance capabilities and represent ends of the performance spectrum.

\begin{table}[t!]
\hspace{2pt}
\footnotesize{}
\centering
\caption{The most pertinent System Under Test (SUT) specifications for the Intel Core-i9 and \ras systems.}
\label{tab:intel-ras-pi}
\resizebox{0.85\columnwidth}{!}{\begin{tabular}{|l|l|l|}
\hline
\textbf{Platform}      & \blue{\textbf{Intel Core-i9}} & \blue{\textbf{\ras}} \\ \hline\hline
CPU Cores              & \blue{18 x-86}         & \blue{4 Arm-A72}                          \\ \hline
CPU Frequency          & 4.2~GHz             & \blue{1.5~GHz}                                      \\ \hline
GPU           & \blue{Nvidia GTX 2080 TI}             & None                                          \\ \hline
Power                  & 350~W       & \textless 1.7~W                                \\ \hline
Cost                  & \$1500      &  \blue{\$75}                               \\ \hline
\end{tabular}
}
\end{table}

Three latencies affect the overall processing time. The first is \textit{t$_{1}$}, which is the latency to extract the state information (Depth Image, RGB Image, etc.) from the server.  The state information is fetched from the server to the SUT. The communication protocol used between the server and the SUT is TCP/IP. Initially, we found that ethernet adapter on Intel Core-i9 faster compared to the ethernet adapter on \ras. We make the \textit{t$_{1}$} latencies between Intel Core-i9 and \ras same by adding artificial sleep for Intel Core-i9 platform.\footnote{The sleep latency value that was added to Intel Core-i9 was determined by doing a ping test with the packet size equal to the size of the data (Depth Image) we fetch from the server and averaged it over 50 iterations.}

The second latency is \textit{t$_{2}$}, which is the policy evaluation time for the SUT (i.e., the Intel Core-i9 or the \ras). The policies are evaluated on the SUT, which predicts the output actions based on the input state information received from the server. The policy architecture used in this work has 40.3 Million (\noobj and \staticobj) and 161.77 Million parameters (\dynamicobj. The \textit{t$_{2}$} latency for \noobj policy on \ras is 396~ms, while on the desktop, equipped with GTX 2080 Ti GPU and Intel Core i9 CPU, it is 11~ms. The desktop is 36$\times$ times faster.

The third latency is \textit{t$_{3}$}. Once the policies are evaluated, it predicts actions. These actions are converted to the low-level actuation using the AirSim flight controller APIs.\footnote{https://github.com/Microsoft/AirSim/blob/master/docs/apis.md} These APIs have a duration parameter that controls the duration of a particular action must be applied. This duration parameter is denoted by \textit{t$_{3}$}, and it is kept the same for both SUTs.

To evaluate the impact of the SUT performance on the overall learning behavior, we keep the \textit{t$_{1}$} and \textit{t$_{3}$} latencies constant for both Intel Core-i9 and \ras systems. We focus only on the difference in the policy evaluation time (i.e., \textit{t$_{2}$}) and study how it affects the overall performance time. Using this setup, we evaluate the best policy determined in Section~\ref{sec:policy-exploration} for environments with no obstacles, static obstacles, and dynamic obstacles.

\subsection{Desktop vs. Embedded SUT Performance}

In \Tab{tab:system-eval}, we compare the performance of the policy on a Intel Core-i9 (high-end desktop) and the \ras. We evaluate the best policy on the \noobj, \staticobj and \dynamicobj environments described previously in Section~\ref{sec:policy-exploration}. 

\begin{table*}[]
\centering
\renewcommand*{\arraystretch}{1.4}
\resizebox{1\columnwidth}{!}{
\begin{tabular}{|l|r|r|r|r|r|r|r|r|r|}
\hline
&\multicolumn{3}{c}{\textbf{No Obstacles}} &\multicolumn{3}{|c}{\textbf{Static Obstacles}}& \multicolumn{3}{|c|}{\textbf{Dynamic Obstacles}}\\ \hline\hline
\textbf{Metric}         & \textbf{Intel Core i9} & \textbf{\ras} & \textbf{Perf. Gap (\%)}& \textbf{Intel Core i9}& \textbf{\ras}&\textbf{Perf. Gap (\%)} & \textbf{Intel Core i9}& \textbf{\ras}&\textbf{Perf. Gap (\%)}\\ \hline
\textit{Inference Latency (ms)} ($\downarrow$)   &   \blue{11.00}                 & \blue{396}             &    \blue{3500}      &   \blue{10}                 & \blue{542.28}             &    \blue{5322.8}  &   \blue{9.3}                 & \blue{948.92}             &    \blue{10103.4}  \\ \hline
\textit{Success Rate (\%)} ($\uparrow$)        &   \blue{91}                & \blue{80.00}             &    \blue{11.00}    &   \blue{84.00}                & \blue{71.00}             &    \blue{13.00}  &   \blue{61.00}                & \blue{55.00}            &    \blue{6.00}            \\ \hline
\textbf{QOF metrics} & \\ \hline 
\textit{Flight Time (s)} ($\downarrow$)   & \blue{ 25.29} & \blue{37.37}  & \blue{47.76}  & \blue{30.258} & \blue{34.44} & \blue{13.85}  & \blue{21.48}  & \blue{35.36} & \blue{64.61}             \\ \hline
\textit{Distance Flown (m)} ($\downarrow$) & \blue{27.59} & \blue{33.06} & \blue{19.82}   & \blue{28.70}    & \blue{32.57} & \blue{13.4} & \blue{23.51} & \blue{32.86} & \blue{39.77}            \\ \hline
\textit{Energy (kJ)} ($\downarrow$)       & \blue{19.68}              & \blue{25.483}        & \blue{29.48}     & \blue{19.2}              & \blue{23.90} & \blue{24.47} & \blue{18.76} & \blue{24.31} & \blue{29.58} \\ \hline 
\end{tabular}
}
\caption{Inference time, success rate, and Quality of Flight (QoF) metrics between Intel Core i9 desktop and \ras for \noobj, \staticobj,  and \dynamicobj. The policy under evaluation is the best policy obtained from policy evaluation (Section~\ref{sec:policy-exploration}). Perf. Gap (\%) is used to quantify the hadware gap.}
\label{tab:system-eval}
\end{table*}
\begin{figure*}[t!]
\centering
\begin{subfigure}{0.32\columnwidth}
  \centering
  \includegraphics[width=1.0\columnwidth]{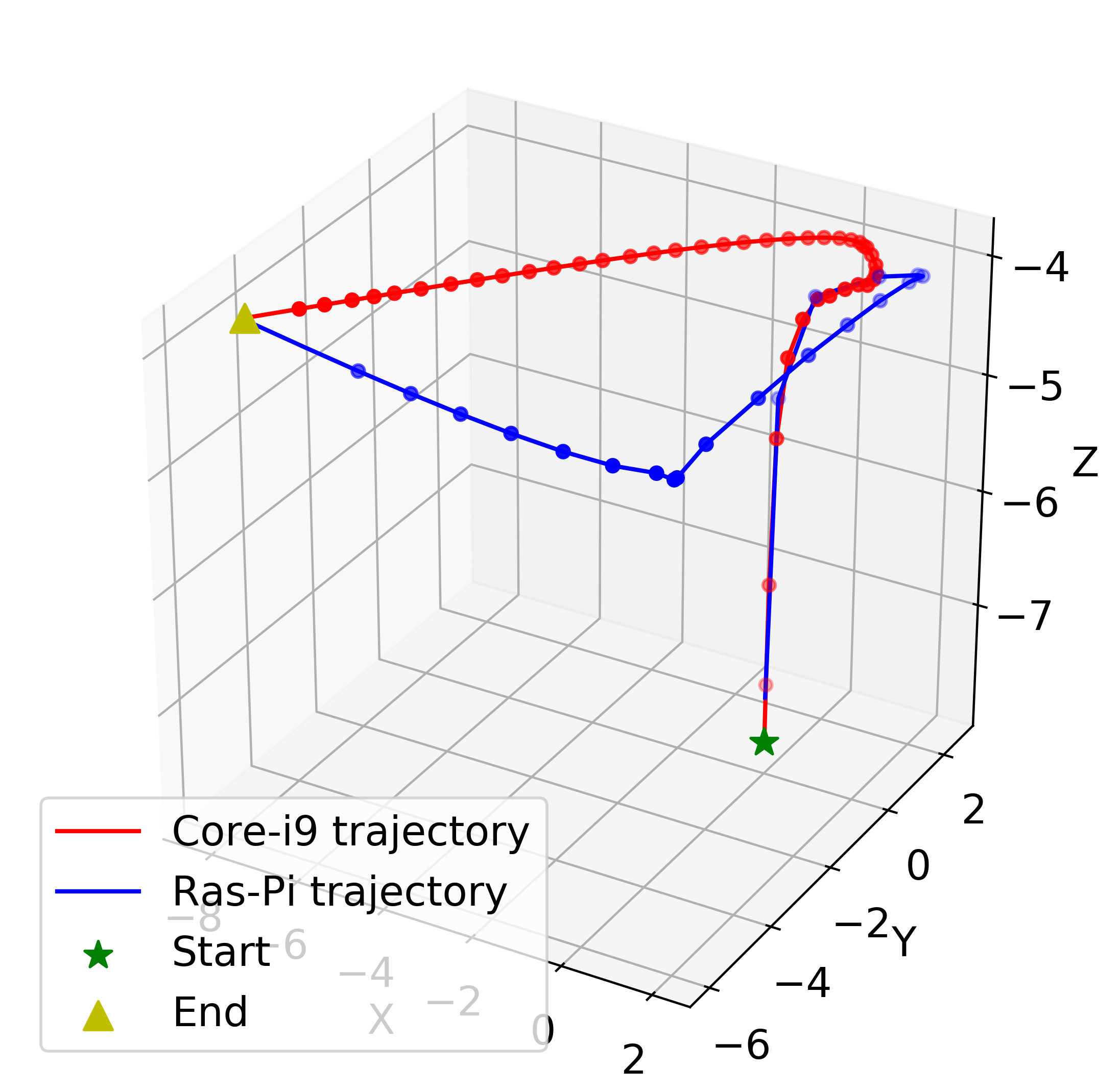}
  \caption{ Trajectory for No Obstacles.}
  \label{fig:env-1-trajectories}
\end{subfigure}
\begin{subfigure}{0.32\columnwidth}
  \centering
  \includegraphics[width=1.0\columnwidth]{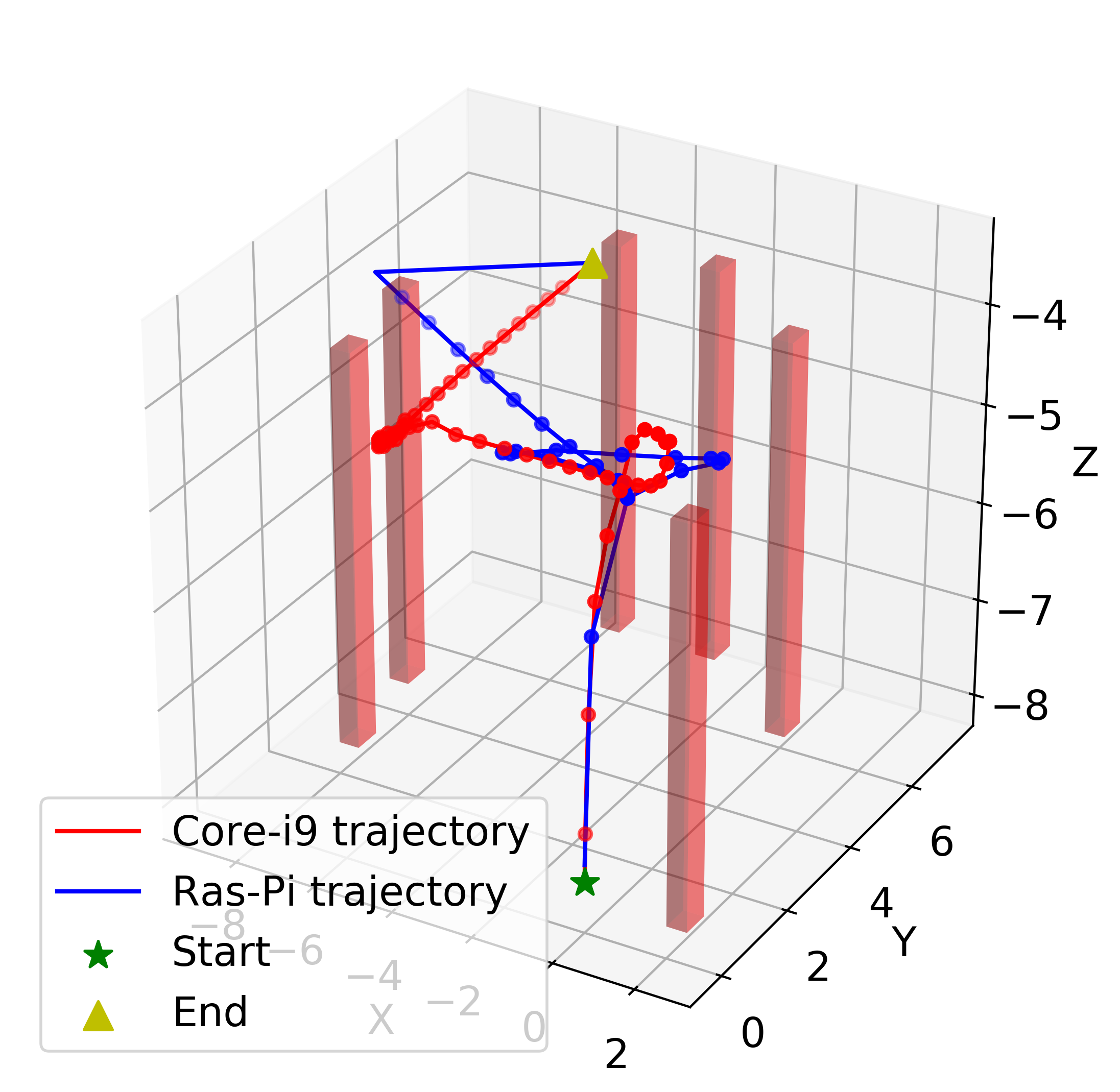}
  \caption{Trajectory for Static Obstacles.}
  \label{fig:env-2-trajectories}
\end{subfigure}
\begin{subfigure}{0.32\columnwidth}
  \centering
  \includegraphics[width=1.0\columnwidth]{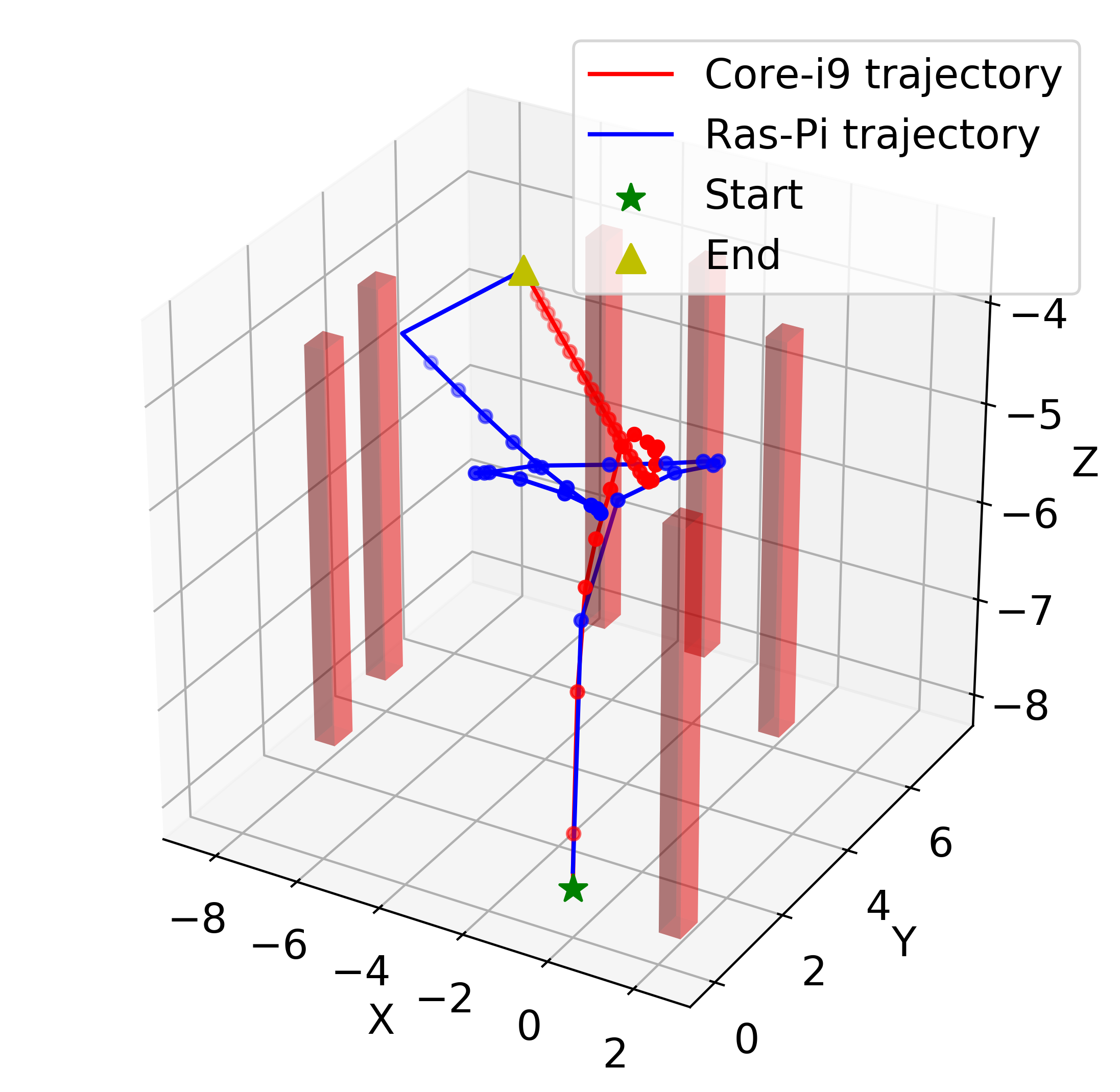}
  \caption{Trajectory for Dynamic Obstacles.}
  \label{fig:env-3-trajectories}
\end{subfigure}
\caption{Figures (a), (b), (c) compare the trajectories of \ras and Intel Core-i9. The red columns in (b) and (c) denotes the position of static obstacles.} 

\label{fig:test}
\end{figure*}

In the \noobj case, the policy running on the high-end desktop is 11\% more successful compared to the policy running on the \ras. The flight time to reach the goal on the desktop is, on average, 25.29~s, whereas on the \ras, it is 37.37~s, which yields a performance gap of around 47.76\%. The distance flown for the same policy on the desktop is 27.59~m, whereas on the \ras, it is 33.06~m, which contributes to a difference of 19.82\%. Finally, the desktop consumes an average of 20~kJ of energy, while the \ras consumes 25.4~kJ, which is 29.48\% more energy.

In the \staticobj case, the policy running on the desktop is 13\% more successful than the policy running on \ras.  The flight time to reach the goal on the high-end desktop is, on average, 30.25~s, whereas on the \ras, it is 34.44~s. That yields a performance gap of around 13.85\%. For the distance flown, the policy running on the desktop has a trajectory length of 28.7~m, whereas the same policy on the \ras has a trajectory length of 32.57~m. This contributes to a difference in 13.4\%. For energy, the policy running on the desktop on an average consumes 19.2~kJ of energy, while policy running on \ras on an average consumes 23.90~KJ of energy, which is about 32\% more energy.

 In the \dynamicobj case, the success rate between the desktop and the \ras is 6\%. The flight time to reach the goal on the desktop is, on average, 21.48~s, whereas on the \ras, it is 35.36~s, yielding a performance gap of around 64.61\%. For the distance flown, the policy running on the desktop has a trajectory length of 23.51~m, whereas the same policy running on \ras has a trajectory length of 32.86~m. This contributes to a difference in 40\%. For energy, the policy running on the desktop, on average, consumes 18.76~kJ of energy while policy running on \ras consumes 24.31~KJ of energy, which is about 30\% more energy.

Overall, across the three different environments, the policy evaluated on the \ras achieves a success rate that is within 13\% compared to the policy assessed on the desktop. While some degradation in performance is expected, the magnitude of the deterioration is more severe for the other QoF metrics, such as flight time, energy, and distance flown. This difference is significant to note because when the policies are ported to resource-constrained compute like the \ras (a proxy for onboard compute in real UAVs), they could perform worse, such as being unable to finish the mission due to low battery.

In summary, the takeaway is that evaluations of policies solely
on a high-end machine do not accurately reflect the real-time performance on an embedded compute system such as those available on UAVs. Hence, relying on success rate as the sole metric is insufficient, though this is by and a large state of the art means to report success. Using Air Learning and its HIL methodology and QoF metrics, we can understand to what extent the choice of onboard compute affects the performance of the algorithm.

\subsection{Root-cause Analysis of SUT Performance Differences} 
\label{sec:understanding-diff}

It is important to understand why the policy performs differently on the Intel~Core~i9 versus the \ras. So, we perform two experiments. First, we plot the policy trajectories on the \ras and compare it to the Intel Core-i9 to understand if there is a flight path difference. Visualizing the trajectories helps us build intuition about the variations between the two platforms. Second, we take an Intel Core-i9 platform and degrade its performance by adding artificial sleep such that the policy evaluation times are similar to that of \ras. This helps us validate if the processing time is giving rise to the QoF metric discrepancy. 

To plot the trajectories, we fix the end goal's position, obstacles, and evaluate 100 trajectories with the same configuration in the \noobj,  \staticobj,  and \dynamicobj environments. The trajectories are shown in \Fig{fig:env-1-trajectories}, \Fig{fig:env-2-trajectories}, and \Fig{fig:env-3-trajectories}. They are representative of repeated trajectories between the start and end goals. The trajectories between the desktop and \ras are very different---the desktop trajectory orients towards the goal and the proceeds directly. The \ras trajectory starts toward the goal, but then drifts, resulting in a longer trajectory. This is likely a result of the actions taken because of stale sensory information, due to the longer inference time; recall there is a 20$\times$ difference in the inference time between the desktop and \ras (Section~\ref{sec:sys-eval-setup} and Table~\ref{tab:system-eval}). 

To further root-cause and test whether the (slower) processing time (\textit{t$_{2}$}) is giving rise to the long trajectories, we take the best performing policy trained on the high-end desktop in the \staticobj environment and gradually degrade the policy's evaluation time by introducing artificial sleep times into the program.\footnote{Adding artificial sleep into the high-end desktop is a simple first-order approximation of the \ras system. In reality, we cannot fully equate the high-end desktop to the \ras since there are other differences (e.g., system architecture, memory sub-system, and power).} Sleep time injection allows us to model the big differences in the behavior of the same policy and its sensitivity to the onboard compute performance.

Table~\ref{tab:system-degrade} shows the effect of degrading the compute performance on policy evaluation. The baseline is the performance on the high-end Intel Core-i9 desktop. Intel Core-i9 (150 ms) and Intel Core-i9 (300 ms) are the scenarios where the performance of Intel Core-i9 is degraded by 150~ms and 300~ms, respectively. As performance deteriorates from 3~ms to 300~ms,  the flight time degrades by 97\%, the trajectory distance degrades by 21\%, and energy degrades by 43\%.

\begin{figure}[h!]
\centering
\includegraphics[width=0.86\columnwidth,keepaspectratio]{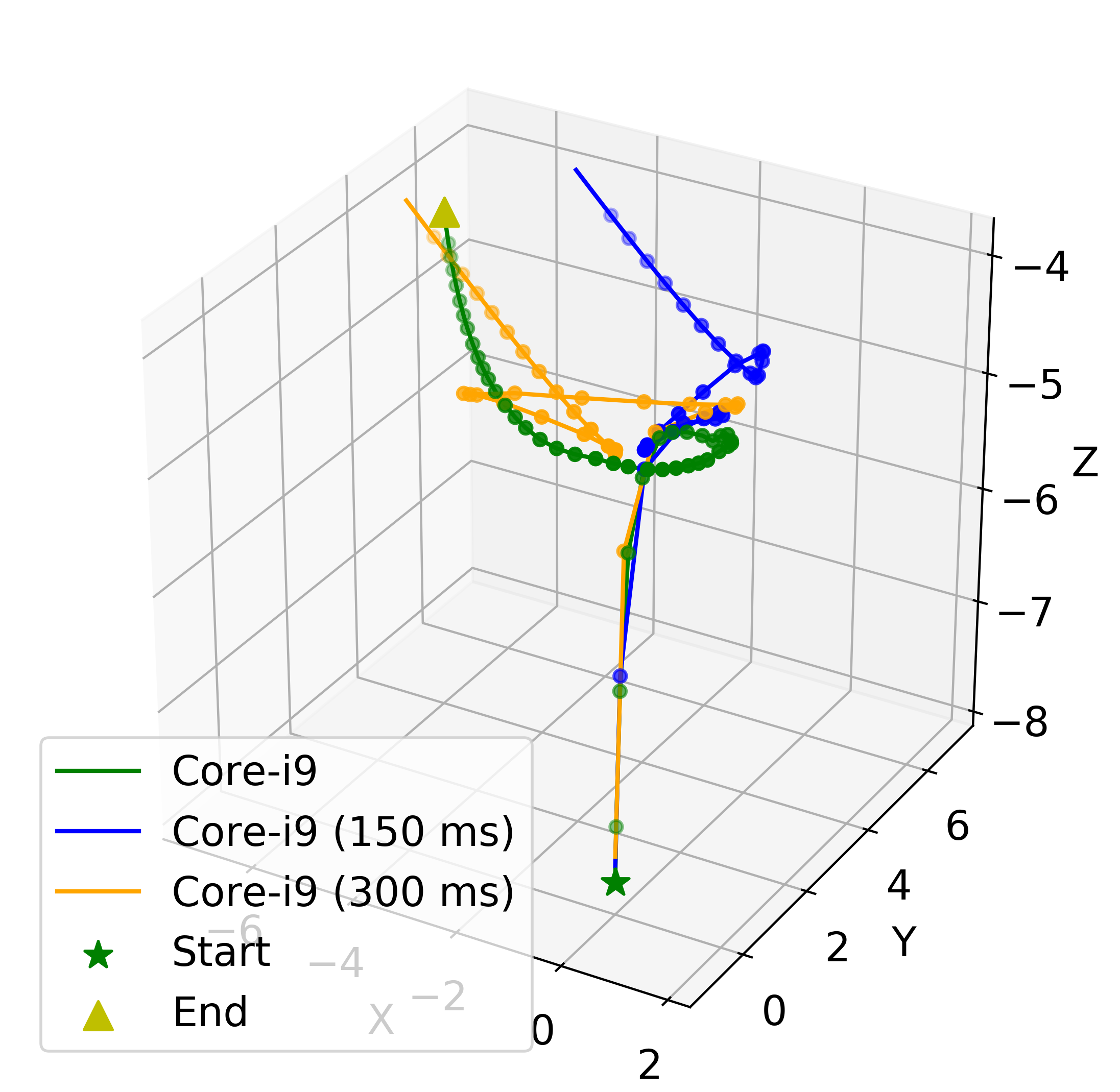}

\caption{Trajectory visualization of  the best-performing policy on Intel Core i9  and artificially degraded versions of Intel Core i9 (150 ms) and Intel Core i9 (300 ms).}
\label{fig:desktop-degradation}
\end{figure}

We visualize degradation impact by plotting the same policy's trajectories on the baseline Intel Core-i9 system and the degraded versions of Intel Core-i9 systems (150 ms and 300 ms). The trajectory results are shown in \Fig{fig:desktop-degradation}. As we artificially degrade, the drift in trajectories gets wider, which increases the trajectory length to reach the goal position, thus degrading the QoF metrics. We also see that the trajectory of the degraded Intel Core i9 closely resembles the \ras trajectory.

In summary, the onboard compute choice and algorithm profoundly affect the resulting UAV behavior and shape of the trajectory. Additional quality of flight metrics (energy, distance, etc.) captures the differences better than just the success rate. Moreover, evaluations done purely on a high-end desktop might show lower energy consumption in a mission, but when the solution is ported to real robots, the solution might consume more energy due to the sub-par performance of the onboard compute. Using the hardware-in-the-loop (HIL) methodology allows us to identify these differences and other performance bottlenecks that arise due to the onboard compute without having to port things to the real robots. Hence, a tool like Air Learning with its HIL methodology helps identify such differences at the early stage.

In the next section, we show how Air Learning HIL can mitigate the hardware gap and characterize the end-to-end learning algorithms and model these characteristics to create robust and performance-aware policies.

\begin{table}[]
\centering
\resizebox{1\columnwidth}{!}{
\renewcommand*{\arraystretch}{1.05}
\begin{tabular}{|l|r|r|r|}
\hline \textbf{Metric}         & \blue{\textbf{Core i9}} & \blue{\textbf{Core i9 (150 ms)}} & \blue{\textbf{Core i9 (300 ms)}} \\ \hline \hline
\textbf{\textit{Inference Latency (ms) ($\downarrow$)}}    &   \blue{11.00}                 & \blue{150.00} & \blue{300.00}   \\ \hline
\textbf{QoF metrics} & \\ \hline
\textbf{\textit{Flight Time (s) ($\downarrow)$ }}    & \blue{24.08}                  & \blue{32.38}             & \blue{47.59}  \\ \hline
\textbf{\textit{Distance Flown (m) ($\downarrow)$}} & \blue{25.64}                  & \blue{28.27}             & \blue{31.11} \\ \hline
\textbf{\textit{Energy (kJ) ($\downarrow$)}}        & \blue{19.09}               & \blue{23.18}          & \blue{27.28} \\ \hline
\end{tabular}
}
\caption{Degradation in policy evaluation using artificially injected  program sleep (proxy for performance degradation). The policy is the best performing policy trained on `Static Obstacles.' The baseline is Intel Core i9 without any artificial sleep. Intel Core i9 (150 ms latency) and Intel Core i9 (300 ms latency) represent scenarios where 150~ms and 300~ms of artificially injected delay added to the policy evaluation.}
\label{tab:system-degrade}

\end{table}

%% file: mitigation.tex
\section{Mitigating the Hardware Gap}
\label{sec:mitigation}

\begin{figure*}[t!]
\hspace{-20pt}
        \begin{subfigure}{.4\linewidth}
        \includegraphics[width=\textwidth]{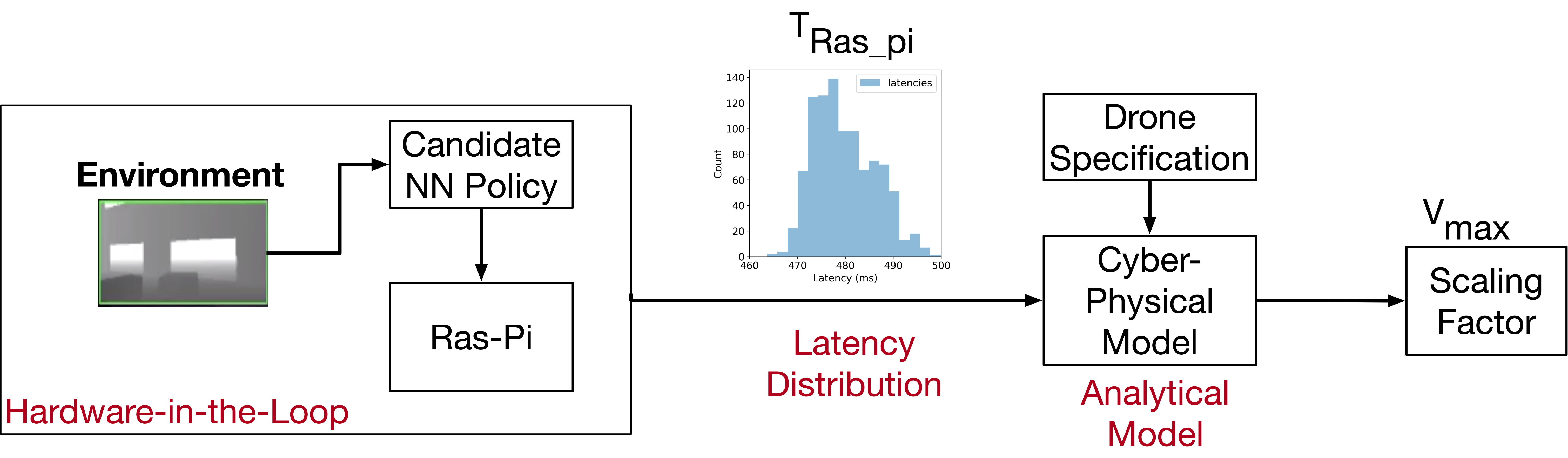}
        \caption{\footnotesize Phase 1: Estimating the latency distribution using HIL.}
        \label{fig:p1}
    \end{subfigure}
    \hspace{10pt}
    \begin{subfigure}{.25\linewidth}
        \includegraphics[width=\textwidth]{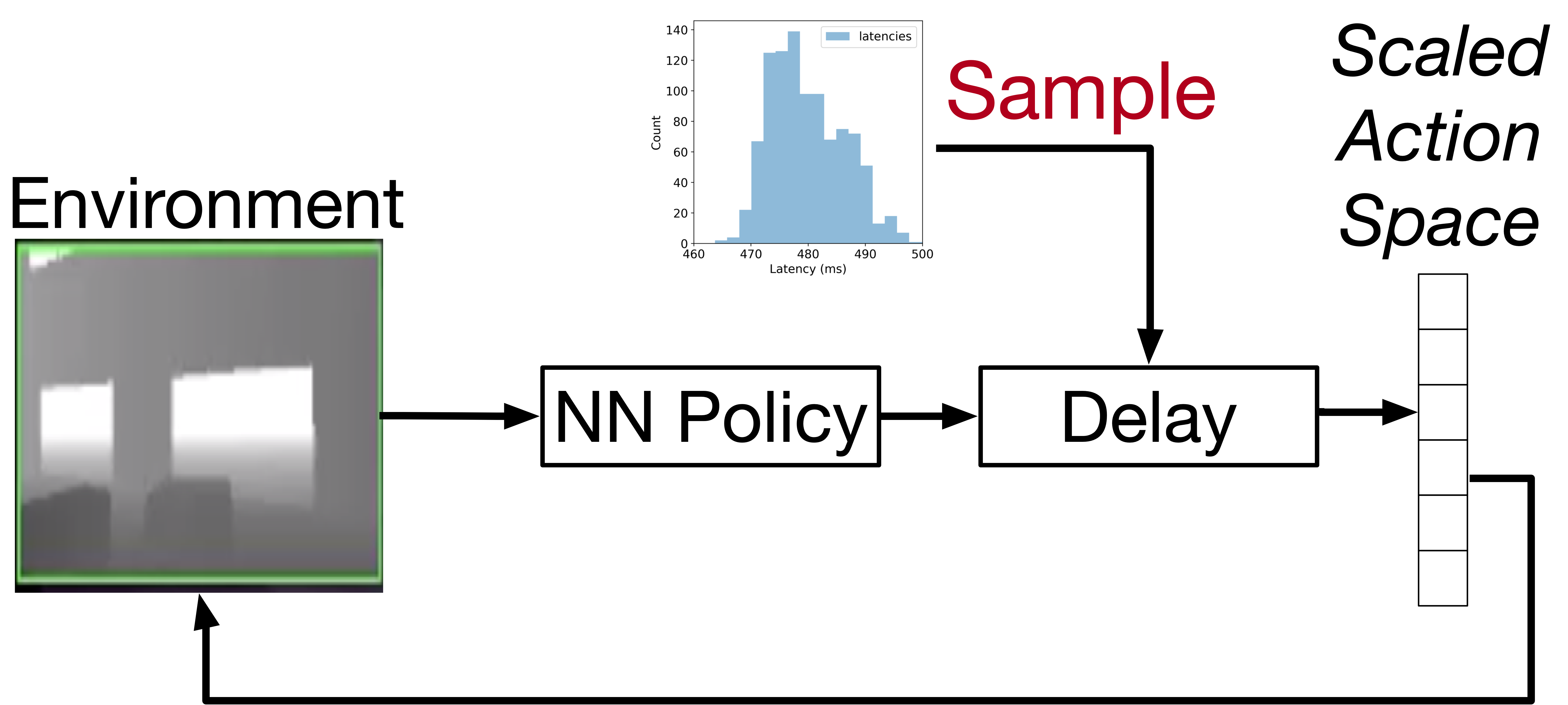}
        \caption{\footnotesize Phase 2: Training with sampling latency distribution.}
        \label{fig:p2}
    \end{subfigure}
    \hspace{10pt}
    \begin{subfigure}{.25\linewidth}
        \includegraphics[width=\textwidth]{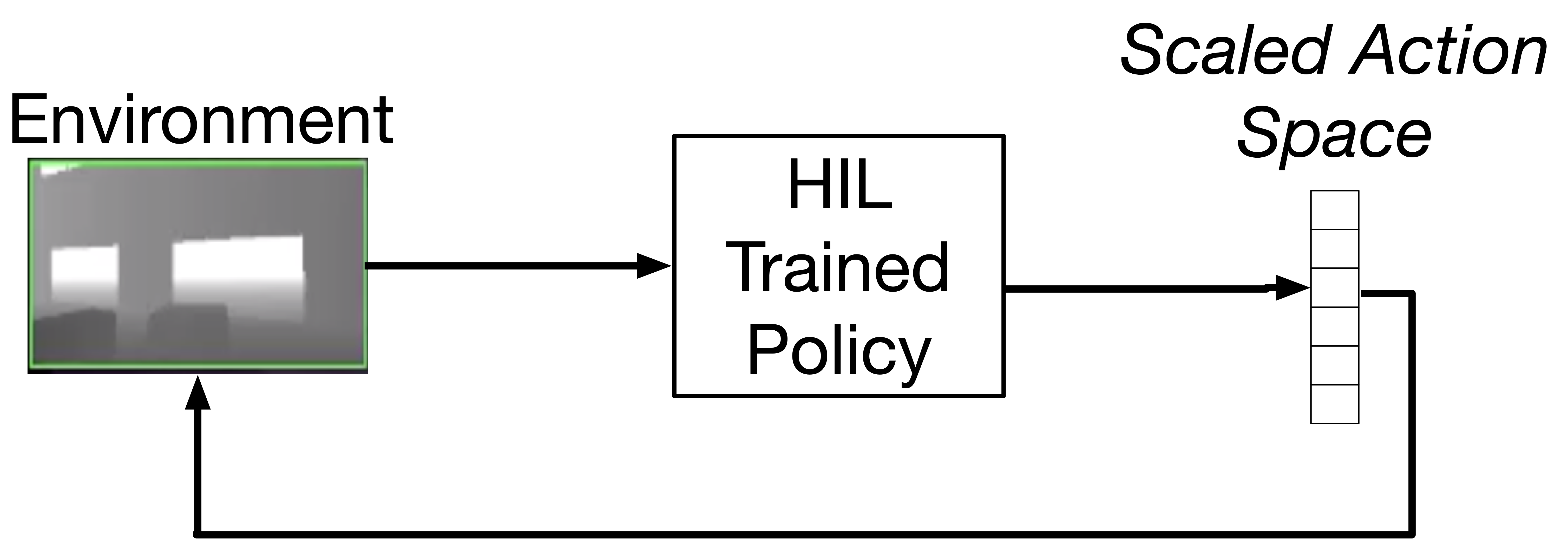}
        \caption{\footnotesize Phase 3: Deploying the policy.}
        \label{fig:p3}
    \end{subfigure}
    \caption{A three-phase methodology for mitigating the hardware gap using hardware-in-the-loop training. (a) In phase 1, we use the hardware-in-the-loop methodology on a candidate policy to get the policy's latency distribution on target hardware (\ras). We use prior work~\cite{f1-roofline} as the cyber-physical model to determine the upper bound for maximum velocity. (b) In phase 2, we use the latency distribution to randomly sample the delay that needs to be added to the policy's training. (c)In phase 3, the HIL trained policy is deployed on the target hardware for evaluation.}
    \label{fig:methodology}
   
\end{figure*}
In this section, we demonstrate how Air Learning HIL technique can be used to minimize the hardware gap due to differences in the training hardware and deployment hardware (onboard compute). To that end, we propose a general methodology where we train a policy on the high-end machine with added latencies to mimic the onboard compute's performance. Using this method, we show that it minimizes the hardware gap from 38\% to less than 0.5\% on flight time metric, 16.03\% to 1.37\% on the trajectory length metric, and 15.49\% to 0.1\% on the energy of flight metric.

One way to mitigate the hardware gap is to directly train the policy on the onboard computer available in the robot~\cite{qt-opt,rl-on-device-1,robel}. Though on-device RL training is practical for ground-based or fixed robots to overcome the `sim2real gap'~\cite{sim2real-1,sim2real-2}, in the context of UAVs, training the RL policy on-device during flight has logistical limitations and not scalable (As explained in Section~\ref{sec:challenges}). Moreover, some of the onboard computers on these UAVs don't have the necessary hardware resources required for on-device RL training. For instance, most hobbyist drones and research UAV platforms (e.g., CrazyFlie) are typically powered by microcontrollers and have a total of 1~MB memory. For most vision based navigation, its storage space is insufficient for the policy weights. Hence these resource constraints make RL training on-device extremely challenging.

To overcome these resource constraint limitations and enable on-device RL training for UAVs, we introduce a methodology that uses HIL for training the RL policy. This methodology allows us to train the RL policy on a high-end machine (e.g., Intel Core-i9 with GPUs) while capturing the latencies incurred in processing the policy in the onboard computer. We describe the details of the methodology below.

\subsection{Methodology}

The methodology is divided into three phases, namely `Phase 1', `Phase 2', and `Phase 3' as shown in \Fig{fig:methodology}. In Phase 1 (\Fig{fig:p1}), we use the HIL to determine three specific latencies namely t$_{1}$, t$_{2}$, and t$_{3}$ defined in Section~\ref{sec:sys-eval-setup}. We capture the latency distribution when the policies are run on-device (e.g., Ras-Pi). The distribution captures the variation in the decision-making times when the policy is deployed in the onboard computer.

Once the latency distribution is captured, we calculate the maximum achievable velocities for safe navigation based on the decision-making time~\cite{high-speed}. This is to ensure that the drone can navigate safely without colliding with an obstacle. We evaluate the maximum safe velocity the aerial robot can travel based on the visual performance model proposed in this work~\cite{f1-roofline}. The model considers the time to action latency, drones' physics (e.g., thrust-to-weight ratio, sensing distance, etc.) to determine the drone's maximum safe velocity.

In phase 2 (\Fig{fig:p2}), we train the policy by adding extra delays sampled from the latency distribution determined in phase 1. The decision-making loop's added delays mimic the typical processing delay when the policy is deployed on the resource-constrained onboard computer. The policy's action space is also scaled based on the maximum velocity achievable based on the decision-making time~\cite{f1-roofline,high-speed}.

Once the policy is trained, in phase 3 (\Fig{fig:p3}), we deployed the trained policy on the onboard compute (Ras-Pi) and evaluate its performance and quality of flight metrics.

\subsection{Experimental Setup and Evaluation}
To validate the methodology, we train a policy on the Static Obstacle environment with at most two to three obstacles. The candidate architecture policy is 5 Layers with 32 Filters based on the template defined in \Fig{fig:policy-eval}.

We use the HIL setup described in \Fig{fig:hil-sys-eval} to evaluate the decision making latency on \ras, which is our target resource-constrained hardware platform. The simulation environment is rendered on the Intel Core-i9 server.  We deploy a randomly initialized policy on the \ras at this stage to benchmark the latencies. We do a rollout of 1000 steps using HIL to capture the variations in decision-making times.

On the high-end server (Intel core-i9 with GTX 2080 TI), we train the candidate policy for the same task (i.e., Static Obstacles) with added delay element in the decision-making loop. The delay element's actual value is randomly sampled from the latency distribution obtained for the candidate policy (5 Layers with 32 Filters) running on the \ras. Also, based on the maximum value of the latency from the distribution, we estimate the upper limit for the safe velocity for drone~\cite{high-speed, f1-roofline}. This upper limit in safe velocity is then used to scale the action space such that at any point, the drone's velocity at each step does not exceed the maximum safe velocity.

\begin{figure}[h!]
\centering
\includegraphics[width=0.7\columnwidth,keepaspectratio]{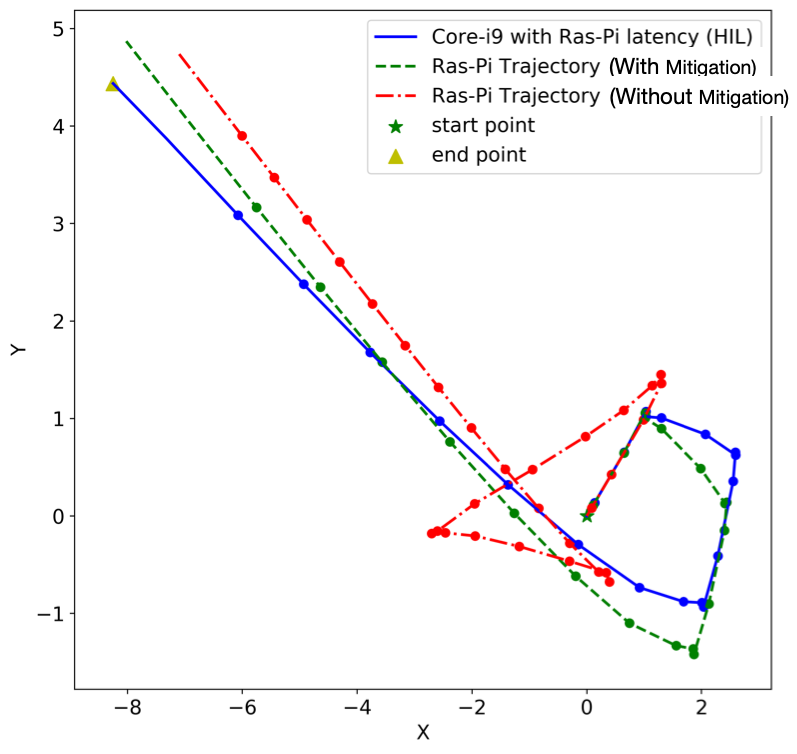}
\caption{Comparison of trajectory for a policy that uses mitigation technique (denoted by the label \blue{``With mitigation''}) with the policy that does not use mitigation technique (represented by the label \blue{``Without mitigation''}). The policy trained on the training machine (denoted by the label (``HIL'') is also plotted for comparison. Using the mitigation technique, we reduced the trajectory length \blue{degradation} from \blue{34.15 m to 29.03m (to within 1.37\%).}}

\label{fig:mitigation}
\end{figure}

Once the candidate policy's training with added latency is complete, we deploy the policy on the \ras platform (target resource-constrained onboard compute). We use the HIL methodology to evaluate the quality of flight metrics on \ras. The comparison in trajectories between Core-i9 and \ras is shown in \Fig{fig:mitigation}. The two trajectories are very similar to each other and do not suffer from larger drifts that were seen before. Table~\ref{tab:mitigation} compares the quality of flight metric. The performance gap (denoted by ``Perf Gap'') is reduced from 38\% to less than 0.5\% on the flight time metric, 16.03\% to 1.37\% on the trajectory length metric, and 15.49\% to 0.1\% on the energy of flight metric.

\begin{table}[]
\resizebox{1\columnwidth}{!}{
\renewcommand*{\arraystretch}{2}
\Huge
\begin{tabular}{|c|c|c|c|c|c|c|}
\hline
\textbf{Metric} &
  \blue{\textbf{\begin{tabular}[c]{@{}c@{}}Intel Core i9 \\ (Without mitigation)\end{tabular}}} &
  \textbf{\begin{tabular}[c]{@{}c@{}}Intel Core i9 \\ (Training Machine)\end{tabular}} &
  \textbf{\begin{tabular}[c]{@{}c@{}}Ras-Pi 4 \\ (With mitigation)\end{tabular}} &
  \textbf{\begin{tabular}[c]{@{}c@{}}Ras-Pi \\ (Without mitigation)\end{tabular}} &
  \textbf{\begin{tabular}[c]{@{}c@{}}Perf Gap\\  (With mitigation)\end{tabular}} &
  \textbf{\begin{tabular}[c]{@{}c@{}}Perf Gap \\ (Without mitigation)\end{tabular}} \\ \hline
\textit{\textbf{Inference Latency (ms)}} & \blue{11} & 400   & 396    & 396   & 1    & 1     \\ \hline
\textit{\textbf{Success Rate}}           & \blue{85} & 85    & 83     & 73    & 2.4  & 16.43 \\ \hline
\multicolumn{6}{|c|}{\textbf{Quatlity of Flight}}                                \\ \hline
\textit{\textbf{Flight Time (s)}}      & \blue{24.08} & 32.62 & 32.44  & 44.93 & 0.5  & 37.73 \\ \hline
\textit{\textbf{Distance Flown (m)}}   & \blue{25.64} & 29.43 & 29.031 & 34.15 & 1.37 & 16.03 \\ \hline
\textit{\textbf{Energy (kJ)}}         & \blue{19.09}  & 24.59 & 24.57  & 28.40 & 0.1  & 15.49 \\ \hline
\end{tabular}}
\caption{Evaluation of quality of flight between \ras and Intel Core-i9 with and without mitigation. After using the methodology to minimize the hardware gap, we were able to reduce the gap from 38\% to less than 0.5\% on the flight time, 16.03\% to 1.37\% on the trajectory length, and 15.49\% to 0.1\% on the energy of flight.}
\label{tab:mitigation}
\end{table}

In summary, we show that training policy with added delay to mimic the target platform can be used to minimize the hardware gap and performance difference between the training machine and the resource-constrained onboard compute.

%% file: future-work.tex
\section{Future Work}

The Air Learning toolset and benchmark that we built can be used for solving several open problems related to UAVs which spans multiple disciplines. The goal of this work was to demonstrate the breadth of \airl as an interdisciplinary tool. In the future, \airl can be used to address numerous other questions, including but not limited to the following.

\textbf{Environments:}
In this work, we focus primarily on UAV navigation for indoor applications~\cite{uav-indoor}. Future work can extend \airl's environment generator to explore new robust reinforcement learning policies for UAV control under harsh environmental conditions. For instance, AirSim weather APIs can be coupled with \airl environment generator to explore new reinforcement learning algorithms for UAV control with different weather conditions.\footnote{AirSim plugin weather APIs can be found here: \url{https://github.com/microsoft/AirSim/blob/master/PythonClient/computer_vision/weather.py}}

\textbf{Algorithm Design:}
Reinforcement algorithms are susceptible to hyperparameter tuning, policy architecture, and reward function. Future work could involve using techniques such as AutoML~\cite{auto-ml} and AutoRL~\cite{auto-rl} to determine the best hyperparameters, and explore new policy architectures for different UAV tasks.

\textbf{Policy Exploration:}
We designed a simple multi-modal policy and kept the policy architecture same across DQN and PPO agent. In future work, one could explore other types of policy architectures, such as LSTM~\cite{rl-lstm} and recurrent reinforcement learning~\cite{r-rl}. 
Another future work could expand our work by exploring energy efficient policies by using the capability available in \airl to monitor energy consumption continuously. Energy-aware policies can be associated with open problems in mobile robots,  such as charging station problem~\cite{kundu2018charging}.

\textbf{System Optimization Studies:} Future work on the system optimization can be classified into two categories. First, one can perform a thorough workload characterization for reducing the reinforcement learning training time. System optimizations will speed up the training process, thus allowing us to build more complex policies and strategies~\cite{dota-5} for solving open problems in UAVs. Second, path to building custom hardware accelerators to improve the onboard compute performance can be explored. Having specialized hardware onboard would allow better real-time performance for UAVs.

%% file: conclusion.tex
\section{Conclusion}
\label{sec:conclusion}
We present \airl, a Deep RL gym and cross-disciplinary toolset, which enables Deep RL research for resource constraint systems, and an end-to-end holistic applied RL research for autonomous aerial vehicles. We use Air Learning to compare the performance of two reinforcement learning algorithms namely DQN and PPO on a configurable environment with varying static and dynamic obstacles. We show that for an end to end autonomous navigation task, DQN performs better than PPO for a fixed observation inputs, policy architecture and reward function. We show that the curriculum learning based DQN agent has a better success rate compared to non-curriculum learning based DQN agent with the same number of experience (steps). We then use the best policy trained using curriculum learning and expose the difference in the behavior of aerial robot by quantifying the performance of the policy using HIL methodology on a resource-constrained Ras-Pi~4. We evaluate the performance of the best policy using quality of flight metrics such as flight time, energy consumed and total distance traveled. We show that there is a non-trivial behavior change and up to 40\% difference in the performance of policy evaluated in high-end desktop and resource-constrained Ras-Pi~4. We also artificially degrade the performance of the high-end desktop where we trained the policy. We observe a similar variation in the trajectory as well as other QoF metrics as observed in Ras-Pi~4 thereby showing how the onboard compute performance can affect the behavior of policies when ported to real UAVs. We also show the impact of energy QoF on the success rate of the mission. Finally, we propose a mitigation technique using the HIL technique that minimizes the hardware gap from 38\% to less than 0.5\% on the flight time metric, 16.03\% to 1.37\% on the trajectory length metric, and 15.49\% to 0.1\% on the energy of flight metric.

%% file: ack.tex
\section*{Acknowledgements}

The effort at Harvard University and The University of Texas at Austin was sponsored by support from Intel.

%% file: appendix.tex
\renewcommand{\thesubsection}{\Alph{subsection}}
\section*{\centering Appendix}
\subsection{Air Learning Environment Generator Knobs}
\label{sec:env-knobs}
Here we list the parameters available in Air Learning environment generator in detail. These parameters are exposed as a game configuration file which can be modified by the end user during runtime. These parameters can also be part of the Deep RL training setup where it can be changed before the onset of new episode.\footnote{\url{https://bit.ly/38WL2CA}}

\textbf{Arena Size:} The \texttt{Arena Size} is the total volume available in the environment. It is represented by \texttt{[length, width, height]} tuple. A large arena size means the UAV has to cover more distance in reaching the goal which directly impacts its energy and mission success. \Fig{fig:arena} an arena size of \texttt{50~m X 50~m X 5~m}. The arena can be customized by adding materials, which we describe in the ``materials'' section.

\textbf{Wall Color:} The \texttt{Wall Color} parameter can be used to set the wall colors of the arena. The parameter takes \texttt{[R, G, B]} tuple as input. By setting different values of \texttt{[R, G, B]}, any color in the visible spectrum can be applied to the walls. The neural network policies show sensitivity towards different colors~\cite{zf-net} and varying these color during training can help the policy to generalize well.

\textbf{Asset:} An \texttt{Asset} in \airl is a mesh in UE4~\cite{ue4-asset}. Any asset that is available in the project can be used as a static obstacle, dynamic obstacle, or both. At simulation startup, \airl uses these assets as either a static or dynamic obstacle. The number of assets that will be spawned in the arena will be equal to the \texttt{\#Static Obstacle} and \texttt{\#Dynamic Obstacle} parameter. By having the ability to spawn any asset as an obstacle, the UAV agent can generalize to avoid collision with different types of obstacle.~\Fig{fig:assets} shows some of the sample assets used in this work.

\textbf{Number of Obstacles:} The \texttt{\#~Static Obstacles} is a parameter that describes the total number of static objects that is spawned in the environment.~\Fig{fig:arena-obstacles} shows some of the assets used as random obstacles by the environment generator. Using this parameter, we can generate environments ranging from very dense to very sparse obstacles. Depending upon the value of this parameter, the navigation complexity can be easy or difficult. A large number of obstacles increases the collision probability and can be used for stressing the efficacy of reinforcement learning algorithms.

\textbf{Minimum Distance:} The \texttt{Minimum distance} is a parameter that controls the minimum distance between two static objects in the arena. This parameter in conjunction with \texttt{\# Static Obstacles} is what determines congestion.

\textbf{Goal Position:} The \texttt{Goal Position} is a parameter that specifies the destination coordinate that the UAV must reach. The \texttt{Goal Position} coordinates should always be inside the arena, and there is error checking for input errors. Similar to \texttt{\#~Static Obstacles}, it increases task complexity. 

\textbf{Number of Dynamic Obstacles:} The \texttt{\# Dynamic Obstacles} is a parameter that describes the total number of obstacles that can move in the environment.

\textbf{Velocity:} The \texttt{Velocity} parameter is a tuple of the form \texttt{[V$_{min}$, V$_{max}$]} that works with \texttt{\# Dynamic Obstacles}. The environment generator randomly chooses a value from this range for the velocity of a dynamic obstacle. This coupled with the \texttt{\# Dynamic Obstacles} helps control how dynamic and challenging the environment is for the aerial robot.

\textbf{Seed:} The \texttt{Seed} parameter is used for randomizing the different parameters in the environment. By setting the same `Seed' value, we can reproduce (and randomize) the environment (obstacle position, goal position, etc.).

As mentioned previously, there is a second category of parameters that can be configured. These are not included in the configuration file. Instead, they are controlled by putting files into folders. Details about them are as follows.

\textbf{Textures:} A \texttt{Texture} is an image that is used on an UE4 asset~\cite{ue4-textures}. They are mapped to the surfaces of any given asset. At startup, the environment generator applies textures to matching assets. Textures and materials (below point) help the training algorithm capture different object features, which is important to help the algorithm generalize.

\textbf{Materials}: A \texttt{Material} is a UE4 asset~\cite{ue4-materials} that can be applied to meshes to control the visual look of the scene. Material is usually made of multiple textures to create a particular visual effect for the asset. At simulation startup, \airl environment generator applies materials to matching assets. 

Materials can help training algorithms on two fronts. First, neural network policy has a sensitivity to capture various material features in the objects~\cite{zf-net,domain-rand}. For instance, the type of material affects how light interacts with the surface, and as a result, an RL based robot that is relying on images as input can learn different things (and act differently) under different materials and the textures that it observes. Second, they can make it challenging for the algorithms using image-based inputs. For instance, shiny and transparent objects are harder to detect~\cite{transparent-1, transparent-2}.

In summary, \airl's environment generator allows any UE4 asset to be loaded into the project, and provides flexibility in the choice of obstacles, materials, and texture. These features are essential to provide a safe sandbox environment where to train and evaluate various deep reinforcement learning algorithms and policies that can generalize well.

\blue{\subsection{Algorithm Exploration}}
\label{sec:algo-eval}
\blue{We explore two RL algorithm types for end-to-end navigation task in autonomous UAVs. The choice of the seed algorithm we used in this work can be classified into discrete action algorithms and continuous action algorithm. For discrete action reinforcement learning algorithm, we use Deep Q Networks (DQN), and for the continuous action algorithm, we use Proximal Policy Optimization (PPO). For both these algorithms, we keep the observation space, policy architecture and reward structure same and compare agent performance.}

\blue{\subsubsection{Training Methodology}}
\blue{The training methodology, policy architecture, reward function, and action space for PPO and DQN agent with and without curriculum learning is described below.}

\blue{\textbf{Non-Curriculum Learning:}}
\blue{We train the DQN agent and PPO agent on the environment with static obstacles. To determine the baseline performance for both the algorithms, we train each agent to 1~Million steps using non-curriculum learning. For non-curriculum learning, we randomize the position of the goal and obstacles every episode to be anywhere in the arena. Simply put, the entire arena acts like one zone as shown in \Fig{fig:ncr_zone}. The checkpoints are saved every~50000 steps and use the last saved checkpoint after 1~Million steps.}

 \begin{figure*}[t!]
\centering
        \includegraphics[width=0.9\columnwidth,keepaspectratio]{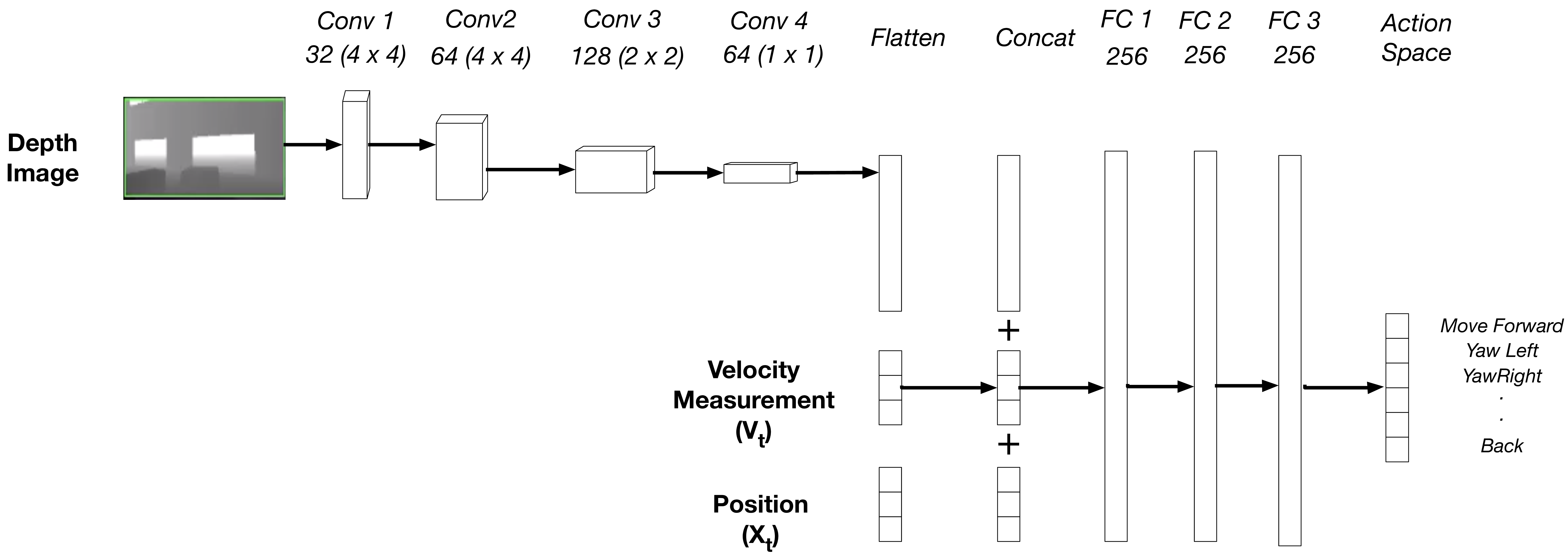}
        
        \caption{\blue{The network architecture for the policy in the PPO and DQN agents. Both the agents take a depth image, velocity vector, and position vector as inputs. The depth image has four layers of convolutions after which the results are concatenated with the velocity and position vectors. In a  \texttt{32 (4 X 4)} convolution filter, \texttt{32} is the depth of the filter and \texttt{(4 X 4)} is the filter size. The combined vector space is applied to the three layers of a fully connected network, each with 256 hidden units. The action space determines the number of hidden units in the last fully connected layer. For DQN agent, we have twenty-five actions, and for PPO agent we have two actions which control the velocity of the UAV agent in X and Y direction.}}

    \label{fig:policy-arch}
\end{figure*}

 \blue{\textbf{Curriculum Learning:} To improve the baseline performance for DQNs and PPO, we employ the curriculum learning~\cite{bengio2009curriculum} approach where the goal position is progressively moved farther away from the starting point of the agent. To implement this, we divide the entire arena into multiple zones namely Zone~0, Zone~1 and Zone~2 as shown in \Fig{fig:cur_zone}. Zone~0 corresponds to the region that is within 16~m from the UAV starting position and Zone~1 and Zone~2  are within 32~m and 48~m respectively. Initially, the position of goal for the UAV is determined randomly such that the goal position lies within Zone~0. Once the UAV agent achieves 50\% success over a rolling window of past 1000 episodes, the position of the goal expands to Zone~1 and so forth. To make sure that the agent does not forget learning in the previous zone, the goal position in the next zone is inclusive of previous zones. We train the agent to progress until Zone~2. Both the agents (PPO and DQN) are trained for 1~Million steps. We checkpoint the policy at every zone so that it can be evaluated on how well it has learned to navigate across all three zones.}
 
\blue{\textbf{Policy Architecture:}}
\blue{The policy architecture for both PPO and DQN agent used is multi-modal in nature. It receives depth image, velocity vector (\textit{V$_t$}) and position vector (\textit{X$_t$}) as inputs as shown in \Fig{fig:policy-arch}. The \textit{V$_t$} is a 1-dimensional vector of the form [\textit{v$_x$}, \textit{v$_y$}, \textit{v$_z$}] where \textit{v$_x$}, \textit{v$_y$}, \textit{v$_z$} are the components of velocity vector in $x$, $y$ and $z$ directions at time \textit{`t'}. The \textit{X$_t$} is a 1-dimensional vector of the form [\textit{X$_{goal}$}, \textit{Y$_{goal}$}, \textit{D$_{goal}$}], where \textit{X$_{goal}$} is the difference in the $x$-coordinate of the goal and $x$-coordinate of the agent's current position, \textit{Y$_{goal}$} is the difference in the $y$-coordinate of the goal and $y$-coordinate of the agent's current position and  \textit{D$_{goal}$} is the euclidean distance to the goal from the agent's current position.}

\blue{The depth image is processed by four convolutions layers whose filter depth and size are \texttt{32 (4 X 4)}, \texttt{64 (4 X 4)}, \texttt{128 (2 X 2)}, and \texttt{64 (1 X 1)} respectively. As an example, in a  \texttt{32 (4 X 4)} filter, \texttt{32} is the depth of the filter and \texttt{(4 X 4)} is the size of the filter. The fourth layer's output is flattened and concatenated with the velocity vector (\textit{V$_t$}) and position vector (\textit{X$_t$}). The combined inputs are then fed to three layers of fully connected layers with \texttt{256} hidden units each. The action space for the agent determines the number of hidden units in the final fully connected layer. For the DQN agent, we have twenty-five discrete actions whereas, for PPO agent, we have two actions. Hence, the final layer for the DQN agent has twenty-five hidden units, and PPO agent has two hidden units. For DQN agent, the activation used for all convolution and the fully connected layer is \textit{ReLU}, and for PPO agent, we use \textit{ReLU} except for the last layer where we use \textit{Tanh} for producing continuous values.}

\blue{\textbf{Action Space:}}
\blue{The action space for DQN consists of twenty-five discrete actions. Out of these twenty-five action spaces, ten actions are for moving forward with different fixed velocities ranging from 1~$m/s$ to 5~$m/s$, five actions are for moving backward, five actions for yawing right with fixed yaw rates of 108~\textdegree, 54~\textdegree, 27~\textdegree, 13.5~\textdegree and 6.75~\textdegree  and another five actions for yawing left with fixed yaw rates of -216~\textdegree, -108~\textdegree, -54~\textdegree, -27~\textdegree and -13.5~\textdegree. At each time step, the policy takes observation space as inputs and outputs one of the twenty-five actions based on the observation. The high-level actions are mapped to low-level flight commands using the flight controller show in Figure~\ref{fig:hil} and as it is implemented.\footnote{https://microsoft.github.io/AirSim/docs/simple\_flight/}}

\blue{The action space for PPO on the other hand consist of velocity components \texttt{v$_x$} (velocity in $x$-direction) and \texttt{v$_y$} (velocity in $y$-direction). At each time step, the policy takes observation as the input and generates continuous values for \texttt{v$_x$} and \texttt{v$_y$}. The values of \texttt{v$_x$} and \texttt{v$_y$} are scaled such that values of the magnitude of velocity lie anywhere between 1~$m/s$ to 5~$m/s$. We use the \texttt{MaxDegreeOfFreedom} option in the AirSim API that calculates the yaw rates automatically to make sure the drone is pointed in the direction it moves.}

\blue{\textbf{Reward:}}
\blue{The reward function for both PPO agent and DQN agent are kept the same and is defined as follows.
\begin{equation}
r =1000*\alpha - 1000*\beta - D_{g} + D_{c}*\gamma
\end{equation}}

\blue{$\alpha$ is a binary variable where `1' denotes if the goal is reached else it is `0'. $\beta$ is also a binary variable where `1' denotes if there is a collision with walls, obstacles or ground else it is `0'. D$_{g}$ is the distance to the goal at any time steps from the agents' current position. If the agent is going away from the goal, the distance to the goal increases thus penalizing the agent. $\gamma$ is also a binary variable which is set to `1' if the agent is closer to the goal. D$_{c}$ is the distance correction which is applied to penalize the agent if it chooses actions which speed up the agent away from the goal. The distance correction term is defined as follows:
\begin{equation}
D_{c} = (V_{max} - V_{now})*t_{max}   
\end{equation}}

\blue{V$_{max}$ is the maximum velocity possible for the agent which for DQN is fixed at 5~$m/s$ and for PPO the outputs are scaled to lie between 1~$m/s$ to 5~$m/s$. V$_{now}$ is the current velocity of the agent and t$_{max}$ is the duration of the actuation.}

\subsection{Policy Architecture vs Runtime Latency Tradeoffs}
Air Learning HIL can also be used to understand the tradeoff between the policy selection and the onboard hardware. In this section, we study the latency tradeoffs for various policies trained for point-to-point navigation policies in No Obstacle, Static Obstacle, and Dynamic Obstacle environments.

\begin{figure}[t!]
\centering
\begin{subfigure}{0.49\linewidth}
        \includegraphics[width=0.98\columnwidth,keepaspectratio]{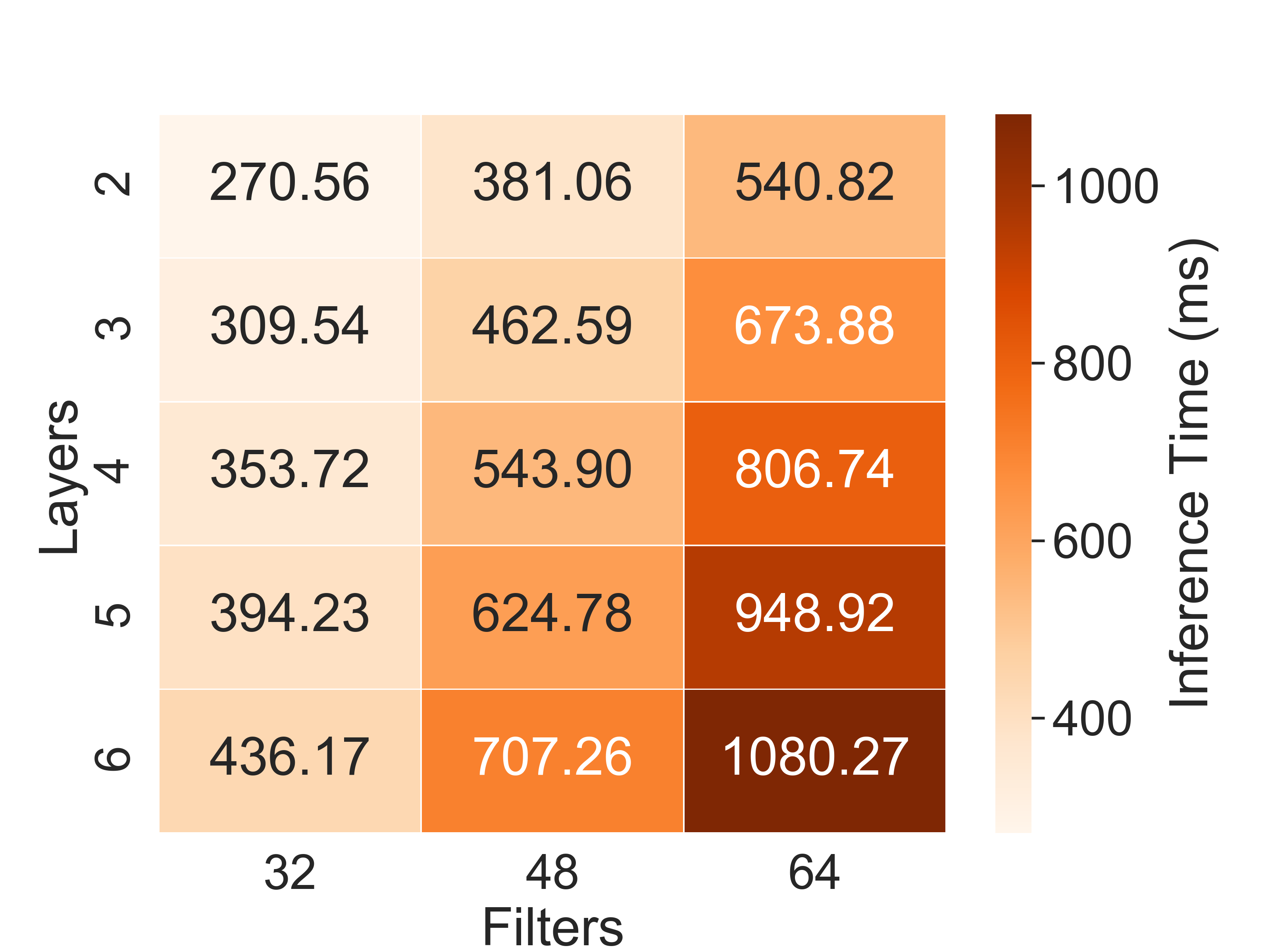}
        \caption{No/Static Obstacle Policies.}
        \label{fig:hw-policy-no_obs}
        \end{subfigure}
        \begin{subfigure}{0.49\linewidth}
        \includegraphics[width=0.98\columnwidth, keepaspectratio]{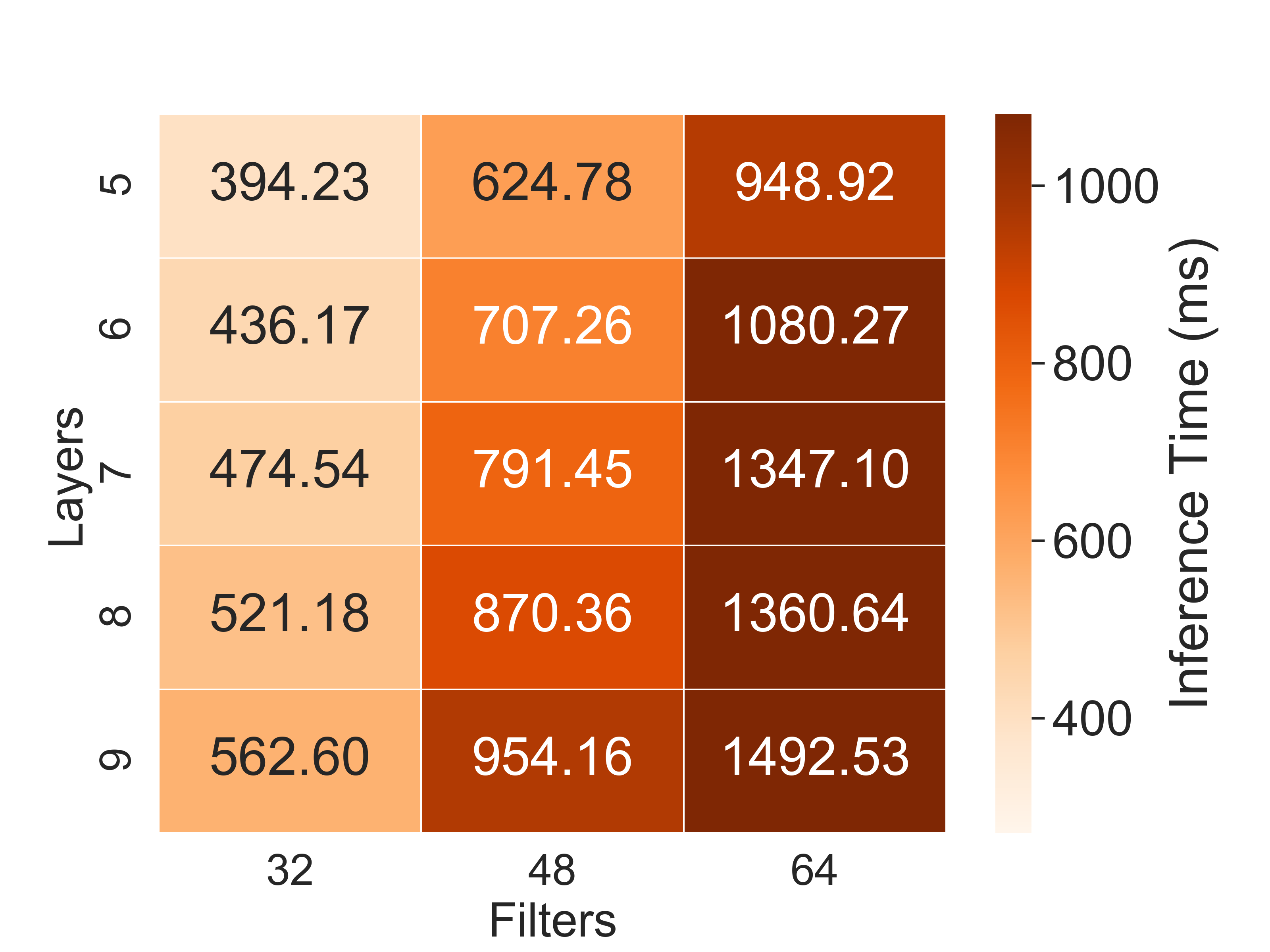}
        \caption{Dynamic Obstacle Policies.}
        \label{fig:hw-policy-dense_obs}
        \end{subfigure}
        \caption{(a) Understanding the trade-offs in latencies between various policy trained for No obstacles and Static obstacles environment. (b) Latencies for various policies trained for Dynamic Obstacles environment. The latencies are averaged over 1000 runs.}
    \label{fig:policy-hw-tradeoff}
\end{figure}

\Fig{fig:policy-hw-tradeoff} shows the latency tradeoff between the size of the policy and the latency to run on Ras-Pi 4. As the policy becomes wider/deeper, we can see that it increases the policy execution time, translating to increased decision making time. Hence while selecting a policy architecture, one must also account for the hardware latency.
 
